\documentclass{article}
\usepackage[utf8]{inputenc} 
\usepackage[T1]{fontenc}    
\usepackage{hyperref}       
\usepackage{url}            
\usepackage{booktabs}       
\usepackage{amsfonts}       
\usepackage{nicefrac}       
\usepackage{microtype}      
\usepackage{graphicx,epsfig,graphics,epsf}
\usepackage{float,amsmath,amsfonts,amssymb,amsthm, bbm}
\usepackage{caption, subcaption}
\usepackage{graphicx}
\usepackage{xcolor}
\usepackage{enumerate}
\usepackage[a4paper,top=3cm,bottom=2cm,left=3cm,right=3cm,marginparwidth=1.75cm]{geometry}
\usepackage{booktabs} 
\usepackage{authblk}
\usepackage{algorithm}
\usepackage{algorithmic}
\usepackage{comment}

\newtheorem{assumption}{Assumption}
\newtheorem{theorem}{Theorem}
\newtheorem{remark}{Remark}
\newtheorem{lemma}{Lemma}
\newtheorem{definition}{Definition}
\newtheorem{cor}{Corollary}
\newtheorem{proposition}{Proposition}

\newcommand{\U}{\mathcal{U}}

\usepackage{hyperref}

\newcommand{\R}{\mathbb{R}}

\newcommand{\Cr}{\mathcal{C}}
\newcommand{\E}{\mathbb{E}}
\newcommand{\D}{\mathcal{D}}

\newcommand{\PP}{\mathbb{P}}
\newcommand{\vecx}{x}
\newcommand{\vecw}{w}

\newcommand{\bigo}{\mathcal{O}}

\newcommand{\W}{\mathcal{W}}

\newcommand{\norm}[1]{\left\|#1\right\|}
\newcommand{\norms}[1]{\|#1\|}
\newcommand{\twonm}[1]{\left\|#1\right\|_2}
\newcommand{\twonms}[1]{\|#1\|_2}

\newcommand{\prob}[1]{\mathbb{P}\left\{#1\right\}}

\newcommand{\EE}{\ensuremath{\mathbb{E}}}

\newcommand{\innerps}[2]{\langle#1,#2\rangle}

\newcommand{\floor}[1]{\lfloor#1\rfloor}

\newcommand{\red}[1]{{\color{red}{#1}}}

\newcommand*\samethanks[1][\value{footnote}]{\footnotemark[#1]}

\begin{document}

\title{Robust Federated Learning in a Heterogeneous Environment}
\date{}
\author[]{Avishek Ghosh\thanks{Equal contributions.}}
\author[]{\hspace{2mm}Justin Hong\samethanks}
\author[]{\hspace{2mm}Dong Yin}
\author[]{Kannan Ramchandran}
\affil[]{\{avishek\_ghosh,~jjhong922,~dongyin,~kannanr\}@berkeley.edu}
\affil[]{Department of Electrical Engineering and Computer Sciences, UC Berkeley}

\maketitle

\begin{abstract}

We study a recently proposed large-scale distributed learning paradigm, namely Federated Learning, where the worker machines are end users' own devices. Statistical and computational challenges arise in Federated Learning particularly in the presence of heterogeneous data distribution (i.e., data points on different devices belong to different distributions signifying different clusters) and Byzantine machines (i.e., machines that may behave abnormally, or even exhibit arbitrary and potentially adversarial behavior). To address the aforementioned challenges, first we propose a general statistical model for this problem which takes both the cluster structure of the users and the Byzantine machines into account. Then, leveraging the statistical model, we solve the robust heterogeneous Federated Learning problem \emph{optimally}; in particular our algorithm matches the lower bound on the estimation error in dimension and the number of data points. Furthermore, as a by-product, we prove statistical guarantees for an outlier-robust clustering algorithm, which can be considered as the Lloyd algorithm with robust estimation. Finally, we show via synthetic as well as real data experiments that the estimation error obtained by our proposed algorithm is significantly better than the non-Byzantine-robust algorithms; in particular, we gain at least by 53\% and 33\% for synthetic and real data experiments, respectively, in typical settings.
\end{abstract}

\section{Introduction}\label{sec:intro}

Distributed computing is becoming increasingly important in many modern data-intensive applications like computer vision, natural language processing and recommendation systems. Federated Learning~(\cite{mcmahan2017federated,mcmahan2016communication,konevcny2016federated}) is one recently proposed distributed computing paradigm that aims to fully utilize on-device machine intelligence---in such systems, data are stored in end users' own devices such as mobile phones and personal computers. Many statistical and computational challenges arise in Federated Learning, due to the highly decentralized system architecture. In this paper, we aim to tackle two challenges in Federated Learning: Byzantine robustness and heterogeneous data distribution.

In Federated Learning, robustness has become one of the major concerns since individual computing units (worker machines) may exhibit abnormal behavior owing to corrupted data, faulty hardware, crashes, unreliable communication channels, stalled computation, or even malicious and coordinated attacks . It is well known that the overall performance of such a system can be arbitrarily skewed even if a single machine behaves in a Byzantine way. Hence it is necessary to develop distributed learning algorithms that are provably robust against Byzantine failures. This is considered in a few recent
works, and much progress has been made~(see \cite{feng2014distributed,blanchard2017byzantine,chen2017distributed,yin2018byzantine,yin2018defending}).

In practice, since worker nodes are end users' personal devices, the issue of data heterogenity naturally arises in Federated Learning. Exploiting data heterogenity is particularly crucial in recommendation systems and personalized advertisement placement, which benefits both the users' and the enterprises. For example, mobile phone users who read news articles may be interested in different categories of news like politics, sports or fashion; advertisement platforms might need to send different categories of ads to different groups of customers. These indicate that leveraging cluster structures among the users is of potential interest---each machine itself may not have enough data and thus we need to better utilize the similarity among the users in the same cluster. This problem has recently received attention in \cite{smith2017federated} in a non-statistical multi-task setting.

We believe that more effort is needed in this area in order to achieve better statistical guarantees and robustness against Byzantine failures. In this paper, we aim to tackle the data heterogeneity and Byzantine-robustness problems simultaneously. We propose a statistical model, along with a $3$ stage algorithm that solves the aforementioned problem yielding an estimation error which is \emph{optimal} in dimension and number of data points. The crux of our approach lies in analyzing a clustering algorithm in the presence of adversarial data points. In particular, we study the classical Lloyd's algorithm augmented with robust estimation. Specifically, we show that the number of misclustered points with the robust Lloyd algorithm decays at an exponential rate when initialized properly. We now summarize the contributions of the paper.

\subsection{Our contributions}

 We propose a general and flexible statistical model and a general algorithmic framework to address the heterogeneous Federated Learning problem in the presence of Byzantine machines. Our algorithmic framework consists of three stages: finding local solutions, performing centralized robust clustering and doing joint robust distributed optimization. The error incurred by our algorithm is optimal in several problem parameters. Furthermore, our framework allows for flexible choices of algorithms in each stage, and can be easily implemented in a modular manner. 

\begin{figure}[t!]
\centering
\begin{tabular}{ccc}
\includegraphics[scale=0.65]{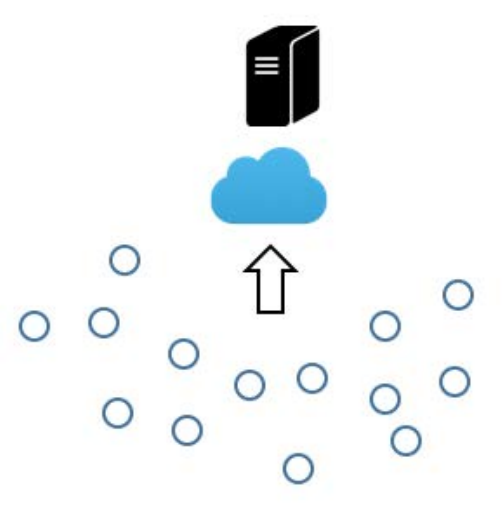}  &
\includegraphics[scale=0.65]{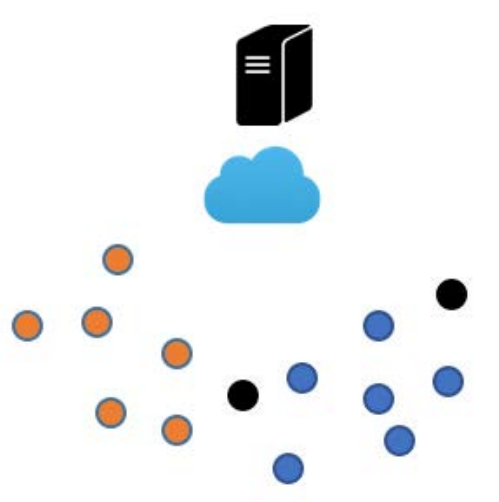} &
\includegraphics[scale=0.65]{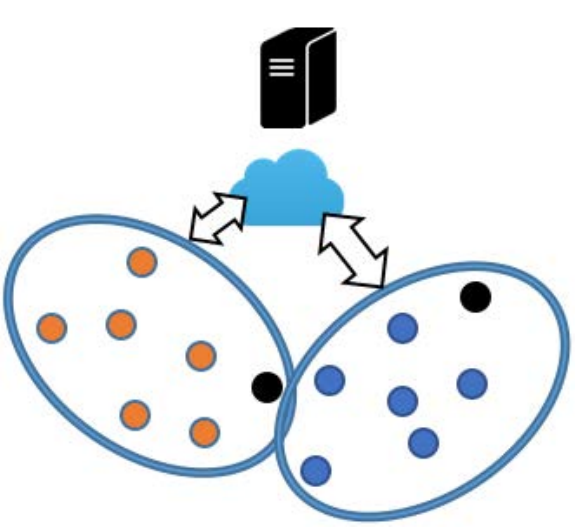}  \\
(i)  &  (ii)  & (iii)
\end{tabular}

\caption{A modular algorithm for Byzantine-robust optimization with heterogeneous data. (i) The $m$ worker machines send their local risk minimizers to the center machine. (ii) The center machine runs robust clustering algorithm (red: cluster $1$, blue: cluster $2$, black: Byzantine machines; center machine may not know which machines are Byzantine). (iii) In each cluster, the center and worker machines jointly run a robust distributed optimization algorithm.}
\label{fig:meta_algo}
\end{figure}

 Moreover, as a by-product, we analyze an outlier-robust clustering scheme, which may be considered as the Lloyd's algorithm with robust estimation. The idea of robustifying the Lloyd's algorithm is not new (e.g. see\cite{har_coresets,charikar_constant} and the references therein) and several robust Lloyd algorithms are empirically well studied. However, to the best of our knowledge, this is the first work that analyzes and prove guarantees for such algorithms in a statistical setting, and might be of independent interest.

 We validate our theoretical results via simulations on both synthetic and real world data. For synthetic experiments, using a mixture of regressions model, we find that our proposed algorithm drastically outperforms the non-Byzantine-robust algorithms. Further, using Yahoo! Learning to Rank dataset, we demonstrate that our proposed algorithm is practical, easy to implement and dominates the standard non-robust algorithms.

\subsection{Related work}\label{sec:references}

\paragraph*{Distributed and Federated Learning:}
Learning with a distributed computing framework has been studied extensively in various settings~\cite{zinkevich2010parallelized,recht2011hogwild,dane,cocoa,yin2017gradient}. Since the paradigm of Federated Learning presented by~\cite{mcmahan2017federated,konevcny2016federated}, several recent works focus on different applications of the problem, such as in deep learning \cite{mcmahan2016communication}, predicting health events from wearable devices, and detecting burglaries in smart homes \cite{pantelopoulos2010survey,Rashidi}.  While \cite{mohri_agnostic} deals with fairness in Federated Learning, \cite{zhao_non-iid,sattler-non-iid} deal with non-iid data. A few recent works study heterogeneity under different setting in Federated Learning, for example see \cite{smith2017federated,zhao2018federated,sahu-heterogeneous,li-hetero} and the references therein. However, neither of these papers explicitly utilize the cluster structure of the problem in the presence of Byzantine machines. Also, in most cases, the objective is to learn a single optimal parameter for the whole problem, instead of learning optimal parameters for each cluster. In contrast, the MOCHA algorithm~\cite{smith2017federated} considers a multi-task learning setting and forms an optimization problem with the correlation matrix of the users being a regularization term. Our work differs from MOCHA since we consider a statistical setting and the Byzantine-robustness.

\paragraph*{Byzantine-robustness:}
The robustness and security issues in distributed learning has received much attention (\cite{alistarh2018sgd,xie2018generalized}). In particular, one recent work by~\cite{li2018rsa} studies the Byzantine-robust distributed learning from heterogeneous datasets. However, the basic goal of this work differs from ours, since we aim to optimize different prediction rules for different users, whereas~\cite{li2018rsa} tries to find a single optimal solution.

\paragraph*{Clustering and mixture models:}
In the centralized setting, outlier-robust clustering and mixture models have been extensively studied. Robust clustering has been studied in many previous works~\cite{chen2008constant,gupta2017local,krishnaswamy2018constant}. One recent work~\cite{charikar} considers a statistical model for robust clustering, similar to ours. However, their algorithm is computationally heavy and hard to implement, whereas the robust clustering algorithm in our paper is more intuitive and straightforward to implement. Our work is also related to learning mixture models, such as mixture of experts~\cite{jacobs1991adaptive} and mixture of regressions~\cite{yi2014alternating,yin2018learning}.

\section{Problem setup}\label{sec:formulation}

 We consider a standard statistical setting of empirical risk minimization (ERM). Our goal is to learn several parametric models by minimizing some (convex) loss functions defined by the data. Suppose we have $m$ compute nodes, $\alpha m~(\alpha < 0.5)$ of which are Byzantine nodes, i.e., nodes that are arbitrarily corrupted by some adversary. Out of the non-Byzantine compute nodes, we assume that there are $K$ different data distributions, $\D_1,\ldots, \D_K$, and that the $(1-\alpha)m$ machines are partitioned into $K$ clusters, $\Cr_1,\ldots,\Cr_K$. Suppose that every node $i \in \Cr_k$ contains $n$ i.i.d.\ data points $\vecx^{i,1},\ldots, \vecx^{i,n}$ drawn from $\D_k$. We also assume that we have no control over the data distribution of the corrupt nodes. Let $f(\vecw;\vecx):\W \rightarrow \R$ be the loss function associated with data point $\vecx$, where $\W\subseteq\R^d$ is the parameter space. Our goal is to find the minimizers of all the $K$ population risk functions. For the $k$-th cluster, the minimizer is $ \vecw^*_k = \arg\min_{\vecw\in\W} F_k(\vecw) (:= \E_{\vecx\sim \D_k}f(\vecw;\vecx))$.

%

The challenges in learning $\{\vecw^*_i \}_{i=1}^{K}$ are: (i) we need a clustering scheme that work in presence of adversaries. Since, we have no control over the corrupted nodes, it is not possible to cluster all the nodes perfectly. Hence we need a robust distributed optimization algorithm. (ii) we want our algorithm to minimize uplink communication cost( \cite{konevcny2016federated}). Throughout, we use $C,C_1,.,c,c_1,.$ for universal constants; whose value may vary from line to line. Also, $||.||$ denotes $\ell_2$ norm.

\section{A modular algorithm for robust Federated Learning in heterogeneous environment}
\label{sec:meta_algo}
In this section, we present a  modular algorithm that consists of $3$ stages---(1) Compute local empirical risk minimizers (ERMs) and send them to the center machine (2) Run outlier-robust clustering algorithm on these local ERMs and (3) Run a communication-efficient, robust, distributed optimization on each cluster (Algorithm~\ref{alg:meta_algo}, also see Figure~\ref{fig:meta_algo}).

\begin{algorithm}[t]
  \caption{A $3$ stage modular algorithm for robust Federated Learning in a heterogeneous environment}
  \begin{algorithmic}[1]
  \REQUIRE Center node, $m$ compute nodes, loss function $f$ .

  \STATE Worker nodes send ERM $\hat{\vecw}^{(i)}:=\mathrm{argmin}_{\vecw \in \mathcal{W}}F^{(i)}(\vecw)$ (for all $i \in [m]$) to the center.
  
  \STATE Center nodes cluster $\{\hat{\vecw}^{(i)}\}_{i=1}^m$ to obtain $\mathcal{C}_1,\ldots,\mathcal{C}_K$.
 
  \STATE At each cluster $\mathcal{C}_1,\ldots,\mathcal{C}_K$ run distributed Byzantine tolerant iterative optimization algorithm.
  \end{algorithmic}\label{alg:meta_algo}
\end{algorithm}

\subsection{Stage I- compute ERMs}
\label{sec:compute_erm_gen}

In this step, each compute node calculates the local empirical risk minimizer (ERM) associated to its risk function send them to the center machine. Since machine $i$ is associated with the local risk function, defined as $F^{(i)}(\vecw)=\frac{1}{n}\sum_{j=1}^{n}f(\vecw,\vecx^{i,j})$, the local ERM, $\hat{\vecw}^{(i)}:=\mathrm{argmin}_{\vecw \in \mathcal{W}}F^{(i)}(\vecw)$. We assume the loss function $f(.;.)$ is convex with respect to its first argument, and so the compute node can run a convex optimization program to solve for $\hat{\vecw}^{(i)}$.

Instead of solving the local risk function directly, the compute node can run an ``online-to-batch conversion'' routine. Each compute node runs an online optimization algorithm like Online Gradient Descent \cite{zinkevich2003online}. At iteration $l$, the compute node picks $\vecw_l$, and incurs a loss of $f(\vecw_l,\vecx_l)$. After $n$ episodes with the sequence of functions $f(.,\vecx_1),\ldots,f(.,\vecx_n)$, the compute node sets the predictor $\bar{\vecw}^{(i)}$ as the average of the online choices $\vecw_1,\ldots,\vecw_n$ made over $n$ instances. This predictor has similar properties like ERMs, however in case of online optimization, there is no need to store all $n$ data points apriori, and the entire operation is in a streaming setup.

\subsection{Stage II- cluster the ERMs}

The second step of the modular algorithms deals with clustering the compute nodes based on their local ERMs. All $m$ compute nodes send local ERMs, $\hat{\vecw}^{(i)}$, for $i \in [m]$ \footnote{For integer $q$, $[q]$ denotes the set of integers $\{1,\ldots,q\}$.} to the center machine, and the center machine runs a clustering algorithm on these data points to find $K$ clusters $\mathcal{C}_1,\ldots,\mathcal{C}_K$. Since compute nodes can be Byzantine, the clustering algorithm should be outlier-robust.

We show (in Section~\ref{sec:naive_cluster}) that if the amount of data in each worker node, $n$ is reasonably large, a simple threshold based clustering rule is sufficient.  This scheme uses the fact that the local ERMs of $2$ machines belonging to a same cluster are close, whereas they are far apart for different clusters. However, if $n$ is small (which is pragmatic in Federated Learning), the aforementioned scheme fails to work.  An alternative is to use a robust version of Lloyd algorithm ($K$-means). In particular: (i) at each iteration, assign the data points to its closest center (ii) compute a robust estimate of the mean with the assigned points for each cluster and use them as new centers and (iii) iterate until convergence.

 The first step is identical to that of the data point assignment of $K$-means algorithm.  There are a few options for robust estimation for mean. Out of them the most common estimates are geometric median \cite{minsker2015}, coordinate-wise median, and trimmed mean . Although these mean estimates are robust, the estimation error $\sim \sqrt{d}$ ($d$ being the dimension) which is prohibitive in large dimension. There is a recent line of work on robust mean estimation that adapts nicely to high dimension \cite{rao_vempala, being_robust_high_d}. In these results, the mean estimation error is either dimension-independent or very weakly dependent on dimension. In Section~\ref{appendix:robust_clustering}, we analyze this clustering scheme rigorously both in moderate and high dimension.

Since we are dealing with the case where $\alpha m$ workers are corrupted, and since we do not have control over the corrupt machines, no clustering algorithm can cluster all the compute nodes correctly, and hence we need a robust optimization algorithm that takes care of the adversarially corrupt (albeit Byzantine) nodes. This is precisely done in the  third stage of the modular algorithm.

\subsection{Stage III- outlier-robust distributed optimization}

After clustering, we run an outlier-robust distributed algorithm on each cluster. Each cluster can be thought of an instance of homogeneous distributed learning problem with possibly Byzantine machines. Hence, we can use the trimmed mean algorithm of \cite{yin2018byzantine} (since it has optimal statistical rate) for low to moderate dimension and the iterative filtering algorithm of \cite{yin2018defending} for high dimension. These algorithms are communication-efficient; the number of parallel iterations needed matches the standard results of gradient descent algorithm.

\section{Main results}
\label{sec:main_results}

 We now present the main results of the paper. Recall the problem set-up of Section~\ref{sec:formulation}. Our goal is to learn the optimal weights $\vecw^*_1,\ldots,\vecw^*_k$. By running the modular algorithm described in the previous section, we compute final output of the learned weights as $\widehat{\vecw}_1,\ldots,\widehat{\vecw}_k$. All the proofs of this section are deferred to Section~\ref{appendix:robust_clustering}. We start with the following set of assumptions.

\begin{assumption}
 The loss function $f(.,\vecx)$ is $G_1$ Lipschitz: $|f(\vecw,\vecx)-f(\vecw_1,\vecx)| \leq G_1 \norm{\vecw-\vecw'}$  for all $ \vecw,\vecw' \in \mathcal{W}$.
\label{ass:second_lipschitz}
\end{assumption}

\begin{assumption}
 $f(.,\vecx)$ is $\lambda$-strongly convex: for all $\vecw$ and $\vecw' \in \mathcal{W}$, 
 \begin{align*}
     \tiny{f(\vecw,\vecx) - f(\vecw',\vecx) - \langle \nabla f(\vecw',\vecx),\vecw-\vecw' \rangle
  \geq \frac{\lambda}{2}\norm{\vecw-\vecw'}^2}
 \end{align*} 
\label{ass:strongly_convex}
\end{assumption}

\begin{assumption}
\label{ass:second_smooth}
$F_k(\vecw)$ is $\lambda_F$ strongly convex, $L_1$ smooth (i.e., $\norms{\nabla^2 F_k(\vecw)}_{\mathsf{op}} \leq L_1 \,\,\,\, \forall \vecw \in \mathcal{W} $).
\end{assumption}

\begin{assumption}
\label{asm:partial}
The function $f(.,\vecx)$ if $L$ smooth. For any $\vecx$ the partial derivative of $f(\vecw,\vecx)$ with respect to the $j$-th coordinate, $\partial_j f(\vecw,\vecx) $ is $L^{(j)}$ Lipschitz and $v$-sub exponential for all $j \in [d]$. 
\end{assumption}

Note that, as illustrated in \cite{yin2018byzantine}, the above structural assumptions on the partial derivative of the loss function are satisfied in several learning problems.

\begin{assumption}
 \label{asm:separation}
 $\{\vecw^*_k\}_{k=1}^K$ are separated: $\min_{i \neq j}\norm{\vecw^*_i - \vecw^*_j} \geq R $ and $n \geq  \frac{G_1^2 L_1 \log m}{\lambda^3} $.
\end{assumption}

\begin{remark}
If $f(\vecw,.)$ is $G_1$ Lipschitz, $\norms{\nabla f(\vecw,.)} \leq G_1$, and hence $G_1^2$ can be $\mathcal{O}(d) $. Also $\lambda$ could be potentially small in many applications. Hence Assumption~\ref{asm:separation} enforces a strict requirement on $n$.
\end{remark}

Let the size of $i$-th cluster is $M_i$ and $\hat{\alpha}_i:=\frac{\alpha m}{M_i + \alpha m}$. Furthermore, let $\max_{i \in [K]} \hat{\alpha}_i  < \frac{1}{2}$.

\begin{theorem}\label{thm:main_result_1}
Suppose Assumptions $1-5$ hold. If Algorithm~\ref{alg:meta_algo} is run with the ``Edge cutting'' (Section~\ref{sec:naive_cluster}) algorithm for stage II and the trimmed mean algorithm (of \cite{yin2018byzantine}) for $T$ iterations with constant step-size of $1/L_1$)  in stage III, then provided $T \geq \widetilde{\bigo}(\frac{L_1 + \lambda_F}{\lambda_F})$, for all $i \in [K]$, we obtain
\begin{align*}
\norms{\widehat{\vecw}_i - \vecw^*_i} \leq \widetilde{\bigo} \left(\frac{\hat{\alpha}_i d }{\sqrt{n}} + \frac{d}{\sqrt{nM_i}} \right).
\end{align*}
with probability
at least $1-m^{-10}-\mathcal{O}\left(\frac{d}{(1+nM_i)^d} \right)$.
\end{theorem}
\begin{remark}
We can remove the assumption of the strong convexity of $f(.,\vecx)$ (Assumption~\ref{ass:strongly_convex}). In that case, under the setting of Theorem~\ref{thm:main_result_1}, for all $i \in [K]$, we obtain
\begin{align*}
    F_i(\widehat{\vecw}_i) - F_i(\vecw^*_i) \leq \widetilde{\bigo} \left(\frac{\hat{\alpha}_i d }{\sqrt{n}} + \frac{d}{\sqrt{nM_i}} \right)
\end{align*}
with high probability.
\end{remark}


\emph{Comparison with an Oracle:} We compare the above bound with an Oracle inequality. We assume that the oracle knows the cluster identity for all the non-Byzantine machines. Since with high probability, the modular algorithm makes no mistake in clustering the non-Byzantine machines, the bound we get perfectly matches the oracle bound.

We now move to the setting where we have no restriction on $n$, and hence $n$ may be potentially much smaller than $d$. This setting is more realistic since data arising from applications (like images and video) are high dimensional, and the amount of data in data owners' device may be small (\cite{mcmahan2017federated}). We start with the following assumption.

\begin{assumption}
\label{asm:subgaussian}
 The empirical risk minimizers, $ \{\hat{\vecw}^{(i)}\}_{i=1}^{(1-\alpha)m}$, corresponding to non-Byzantine machines  are sampled from a mixture of $K$ $\sigma$-sub-gaussian distributions.
 \end{assumption}

We emphasize that several learning problems satisfy Assumption~\ref{asm:subgaussian}. We now exhibit one such setting where the \emph{empirical risk minimizer} is Gaussian. We assume that machine $i$ belongs to cluster $\mathcal{C}_k$. Recall that $\{\vecx^{i,j} \}_{j=1}^{n}$ denote the data points for machine $i$.

\begin{proposition}
\label{prop:sub_Gaussian}
Suppose the data $\{\vecx^{i,j} \}_{j=1}^{n}$ are sampled from a parametric class of generative model: $ \vecx^{i,j} = \langle \chi_j, \vecw^*_k \rangle + \Upsilon_j $ with covariate $\chi_j^T \in \R^d$ and i.i.d noise $\Upsilon_j \sim \mathcal{N}(0,\upsilon^2)$. Then, with quadratic loss, the distribution of the empirical risk minimizer $\hat{\vecw}^{(i)}$ is Gaussian with mean $\vecw^*_k$.
\end{proposition}

 In general, sub-Gaussian distributions form a huge class, including all bounded distributions. For non-Byzantine machines, we assume the observation model: $\hat{\vecw}^{(i)} = \theta_{z_i} + \tau_i $ where $\{z_i\}_{i=1}^{(1-\alpha)m} \in [K]$ are unknown labels and $\{\theta_i\}_{i=1}^{K}$ are the unknown means of the sub-gaussian distribution. We denote $\{\tau_i\}_{i=1}^{(1-\alpha)m}$ as independent and zero mean sub-gaussian noise with parameter $\sigma$. We propose and analyze a robust clustering algorithm presented in Algorithm~\ref{alg:trimmed_k_means}. At iteration $s$, let $z_i^{(s)}$ be the label of the $i$-th data point, and $\hat{\theta}^{(s)}_g$ for $g \in [K]$ be the estimate of the centers.

\begin{algorithm}[h]
  \caption{Trimmed $K$-means}
  \begin{algorithmic}[1]
  \REQUIRE Observations $\hat{\vecw}^{(1)}, \ldots, \hat{\vecw}^{(m)}$, initial labels $\{z_i^{(0)}\}_{i=1}^m$.
  \FOR{$s=1,2, \ldots $}

  \STATE Form $K$ buckets with data points in each bucket having same $z_i^{(s)}$. In each bucket:\\
Compute geometric median of the data points. Construct a ball of radius $C\sigma \sqrt{d}$ around the geometric median (for constant $C$) and compute sample mean of all the data points inside the norm ball as center estimates $\{\hat{\theta}_g^{(s+1)}\}_{g=1}^K$.
  \STATE Re-assign data to the closest center: for all $i \in [m]$, $
    z_i^{(s+1)} = {\mathrm{argmin}_{g \in [K]}} \norms{\hat{\vecw}^{(i)} - \hat{\theta}_g^{(s+1)}}_2$.
  \ENDFOR
  \end{algorithmic}\label{alg:trimmed_k_means}
\end{algorithm}
  In Algorithm~\ref{alg:trimmed_k_means}, we retain the nearest neighbor assignment of the Lloyd algorithm but change the sample mean estimate to a robust mean estimate using geometric median-based trimming.

 We now introduce a few new notations. Let $\Delta := \min_{g\neq h \in [K]} \norms{\theta_g - \theta_h}$ denote the minimum separation between clusters. The worst case error in the centers are determined by $ \Lambda_s = \max_{h\in [K]}\norms{\hat{\theta}_h^{(s)} - \theta_h}/\Delta$. Consequently we define $G_s$ as the maximum fraction of misclustered points in a cluster (maximized over all clusters). In Section~\ref{sec:general_sub_gaussian} and \ref{sec:details_gen_K} (of the supplementary material), these quantities are formally defined along with the initialization condition, $\Lambda_0$.

Recall that $|\mathcal{C}_i|=M_i$ and note that from Theorem~\ref{thm:main_trim}, when Algorithm~\ref{alg:trimmed_k_means} is run for a constant $S$ number of iterations, we get $G_S \leq \varrho$ with high probability. Also, let $\tilde{\alpha}_i = \left ( \frac{\varrho M_i + \alpha m}{M_i + \alpha m} \right )$. Since $G_S$ denotes fraction of non-Byzantine machines that are misclustered, $\tilde{\alpha}_i$ denotes the worst case fraction of Byzantine machines in cluster $i$. We assume $\max_{i \in [K]} \tilde{\alpha}_i < \frac{1}{2}$.

\begin{theorem}\label{thm:main_result_2}
Suppose Assumptions~\ref{ass:strongly_convex}, \ref{ass:second_smooth}, \ref{asm:partial} and \ref{asm:subgaussian} hold along with the separation and initialization conditions (Assumptions~\ref{asm:k_mix} of Section~\ref{sec:robust_lloyd}). Furthermore, suppose Algorithm~\ref{alg:meta_algo} is run with ``Trimmed $K$ means'' (Algorithm~\ref{alg:trimmed_k_means}) for stage II for a constant $S$ iterations; and the trimmed mean algorithm (of \cite{yin2018byzantine}) for stage III for $T$ iterations with constant step-size of $1/L_1$. Then, provided $T \geq \widetilde{\bigo}(\frac{L_1 + \lambda_F}{\lambda_F})$,  for all $i \in [K]$, we have

\begin{align*}
\norms{\hat{\vecw}_i - \vecw^*_i} \leq \widetilde{\bigo} \left(\frac{\tilde{\alpha}_i d }{\sqrt{n}} + \frac{d}{\sqrt{n M_i}} \right).
\end{align*}
with probability
at least $1-m^{-10}-\mathcal{O}\left(\frac{d}{(1+n M_i)^d}\right)$.
\end{theorem}
\begin{remark}
Like before, we can remove Assumption~\ref{ass:strongly_convex} and obtain guarantee on $F_i(.)$ for all $i\in [K]$.
\end{remark}

\emph{Comparison with the oracle:} Recall that the oracle knows the cluster labels of all the non-Byzantine machines. Hence, the worst case fraction of Byzantine machines will be $\hat{\alpha}_i = \frac{\alpha m}{M_i + \alpha m}$. Consequently, we observe that the obliviousness of the clustering identity hurts by a factor of $(\tilde{\alpha}_i - \hat{\alpha}_i)\frac{d}{\sqrt{n}}=  (\frac{\varrho M_i}{M_i + \alpha m} ) \frac{d}{\sqrt{n}}$ in the precision of learning weight $\vecw^*_i$. A few remarks are in order.

\begin{remark}
As seen in Section~\ref{sec:robust_algo}, if $K=2$ and $\theta_1 = -\theta_2$, we show that $\varrho = 0$ if ``Trimmed $K$ means'' is run for at least $3 \log m$ iterations provided $\frac{\Delta}{\sigma} \geq C \sqrt{\log m}$. Hence our precision bound matches perfectly with the oracle bound.
\end{remark}

\begin{remark}
The dependence on $d$ can be improved if \emph{iterative filtering} algorithm (\cite{yin2018defending}) is used in stage III of the modular algorithm. We get $\norms{\hat{\vecw}_i - \vecw^*_i} \leq \widetilde{\bigo} \left( \frac{\sqrt{\tilde{\alpha}_i}}{\sqrt{n}}+\frac{\sqrt{d}}{\sqrt{nM_i}}\right)$ with high probability.
\end{remark}

\subsection{Oracle optimality}
\label{sec:lower_bound}

 In the presence of the oracle, our problem decomposes to $K$ homogeneous ones. We study the dependence of the estimation error of Theorem~\ref{thm:main_result_2} on $n,d, \alpha$, and $M_i$ under such a setting.
 
 \paragraph*{Dependence on $(n,M_i,\alpha)$:}  We compare our results with the lower bounds presented in  \cite[Observation 1]{yin2018byzantine} assuming $d$ is constant. It is immediate that the dependence on $n$ and $M_i$ is optimal. To see the dependence on $\alpha$, we first consider the special case of $K=2$ with centers $\theta_1 = -\theta_2$. Here $\tilde{\alpha}_i= \frac{\alpha m}{M_i + \alpha m}$. Typically, $M_i \gg \alpha m$ and hence $\tilde{\alpha}_i \approx\frac{\alpha m}{M_i}$. Comparing with the bound in \cite[Observation 1]{yin2018byzantine}, the dependence on $\alpha$ is near optimal in this case. However for a $K$ cluster setting, $\tilde{\alpha}_i$ may not be linear in $\alpha$ in general (since $\varrho$ is not proportional to $\tilde{\alpha}_i$).

 \paragraph*{Dependence on dimension $d$:} In this setting, instead of running the trimmed mean algorithm as the distributed optimization subroutine, we run the iterative filtering algorithm of \cite{yin2018defending},  and as shown in Remark 3, the dependence on $d$ when compared with the lower bound of  \cite[Observation 1]{yin2018byzantine} is optimal. Note that in this case, the dependence on $\tilde{\alpha}_i$ becomes sub-optimal.

\section{Robust clustering}
\label{sec:robust_lloyd}

 In Stage II of the modular algorithm, we cluster the local ERMs, $\hat{\vecw}^{(i)}$ in the presence of Byzantine machines. To ease notation, we write $y_i = \hat{\vecw}^{(i)}$. Recall that for non Byzantine data-points, we have $ y_i = \theta_{z_i} + \tau_i $, with unknown labels $\{z_i\}_{i=1}^{(1-\alpha)m} \in [K]$, unknown centers $\{\theta_i\}_{i=1}^{K}$ and $\sigma$ sub-Gaussian noise $\{ \tau_i \}_{i=1}^{(1-\alpha)m}$. For Byzantine data points $y_i$ is arbitrary. It is worth mentioning here that the classical Lloyd can be arbitrarily bad since the adversary may put the data points far away, thus causing the sample mean-based subroutine of the algorithm to fail. As a performance measure, we define the fraction of misclustered non-Byzantine data points at iteration $s$ as, $ A_s = \frac{1}{(1-\alpha)m}\sum_{i \in \mathcal{M}}  I\{z_i^{(s)} \neq z_i \}$, where $\mathcal{M}$ denotes the set of non-Byzantine data points with $|\mathcal{M}|=(1-\alpha)m$. We first concentrate the special case where $K=2$ with centers $\theta^*$ and $-\theta^*$, and hence $ y_i = z_i \theta^* + \tau_i$. With slight abuse in notation, the labels are $z_i \in \{ -1,+1 \}$ and hence, $ z_iy_i = \theta^* +z_i \tau_i = \theta^* +\xi_i$, where $\xi_i \sim \mathcal{N}(0,\sigma^2I_d)$. This can be thought of estimating  $\theta^*$ from samples $z_iy_i$.

\subsection{Symmetric $2$ clusters with Gaussian mixture}
\label{sec:symmetric_cluster}

We analyze Algorithm~\ref{alg:trimmed_k_means} in the above-mentioned setting. The performance depends on the normalized signal-to-noise ratio, $ r:=\norm{\theta^*}/(\sigma\sqrt{1+\eta}) $, where $\eta = 9d/(1-\alpha)m$. At iteration $s$, let $\beta$ be the fraction of data-points being trimmed by Algorithm~\ref{alg:trimmed_k_means} and let $\hat{\theta}^{(s)}$ be the estimate of $\theta^*$.
 
\begin{assumption}
\label{asm:snr}
(i) (SNR) We have $\frac{\norms{\theta^*}}{\sigma} \geq C_{th}$ and $m \geq m_{th}$ (ii) (Initialization) $
A_0 <  \frac{1}{2}-\frac{2.56}{r}-\frac{1}{2\sqrt{(1-\alpha)m}}-\frac{\varepsilon}{2}$, where $m_{th}$, $C_{th}$ are sufficiently large and $\varepsilon$ is sufficiently small constants.
\end{assumption}
Hence we require a constant SNR and $A_0$ needs to be
slightly better than a random guess.

\begin{theorem}
\label{thm:2cluster}
Suppose Assumptions~\ref{asm:subgaussian} and ~\ref{asm:snr} hold. For $\alpha \leq \beta < c/d $ and for $s\geq 0$, $A_s$ satisfies

\begin{equation}
A_{s+1} \leq A_s (A_s + \frac{8}{r^2}) + \frac{2}{r^2} +  \sqrt{4\log((1-\alpha)m)/((1-\alpha)m)} \nonumber
\end{equation}

with probability at least $1- c_1 m^{-3}-c_2m\exp(-d)-2\exp(-\norms{\theta^*}^2/3\sigma^2)$. Furthermore, for $s \geq 3\log m$, $ A_s \leq \exp ( -\norms{\theta^*}^2/16\sigma^2 ) $ with high probability.
\end{theorem}

Hence, if $\norms{\theta^*}/\sigma \gtrsim  \sqrt{\log m}$, then after $3\log m$ steps, $A_s < \frac{1}{(1-\alpha)m}$ implying $A_s=0$, which matches the oracle bound ($\varrho =0$) mentioned after Theorem~\ref{thm:main_result_2}. Also, here we can tolerate  $\alpha \sim 1/d$, which can be prohibitive for large $d$. In the general $K$-cluster case, we improve the tolerance level from $1/d$ to $1/\sqrt{d}$ (Theorem~\ref{theorem:k_mix}), and in Section~\ref{sec:clustering_high_dimension} we completely remove the dependence on $d$.

\subsection{$K$ clusters with sub-Gaussian mixture}
\label{sec:general_sub_gaussian}

 We now analyze the general $K$-cluster setting and with sub-Gaussian noise. The details of this section are deferred to Section~\ref{sec:details_gen_K} of the Appendix. Similar to $A_s$, we define a cluster-wise misclustering fraction $G_s$ and the trimmed cluster-wise misclustering fraction as $G_s^\mathcal{U}$ at iteration $s$. Recall the definition of $\Delta$ and $\Lambda_0$ from Section~\ref{sec:main_results} and denote the minimum cluster size at iteration $s$ as $\gamma_1$. Also define $\alpha_h$ and $\beta_h$ as the fraction of adversaries and trimmed points respectively for the $h$-th cluster. Furthermore, let $\alpha'$ be the maximum adversarial fraction (after trimming) in a cluster and $ r_1 = (\Delta/\sigma)\sqrt{\gamma_1/(1 + \frac{Kd}{(1-\alpha)m})}$ be the normalized SNR.
 
\begin{assumption}
 \label{asm:k_mix}
 We have:  (a) $(1-\alpha)m \gamma_1^2 \geq C_1K\log((1-\alpha)m)$; (b) (SNR) $\Delta \geq C_3 \sigma \sqrt{K} $; and (c) (Initialization) $\Lambda_0 \leq \frac{1}{2}- \frac{4}{\sqrt{r_1}} - \frac{\varsigma}{2}$, for a small constant $\varsigma$.
 \end{assumption}

Hence the separation (of means) is $\mathcal{O}(\sqrt{K})$, which matches the standard separation condition for non-adversarial clustering (\cite{kumar2010clustering}).  Let $\varrho = \Gamma'(c/r_1^2 +\sqrt{\frac{5K\log((1-\alpha)m)}{\gamma_1^2(1-\alpha)m}}) + \rho$, where $\Gamma' = \max_h \frac{1-\beta_h}{1 - \alpha_h}$ and $\rho = \max_h \frac{\beta_h}{1 - \alpha_h}$. We have the following result:
\begin{theorem}
\label{theorem:k_mix}
With Assumption~\ref{asm:k_mix} and $\alpha' \leq c/\sqrt{d}$, the cluster-wise misclustering fraction $G_S^\U$ satisfies

\begin{equation}
G_{s+1}^\U \leq \frac{C_5}{r_1^2} + \frac{2C_2}{r_1^2}G_s^\U + \frac{C_2C_4}{r_1^2}G_s^\U
    + \sqrt{5K\log((1-\alpha)m)/(\gamma_1^2(1-\alpha)m)}\nonumber
\end{equation}
with probability exceeding $1 - 2((1-\alpha)m)^{-3} - \exp(-0.3(1-\alpha)m) - \exp(-0.5(1-\alpha)m)$. Furthermore, if Algorithm~\ref{alg:trimmed_k_means} is run for a constant $S$ iterations, $G_S \leq \varrho$ with high probability.
\end{theorem}

\subsection{Robust clustering in high dimension}
\label{sec:clustering_high_dimension}

In Sections~\ref{sec:symmetric_cluster} and~\ref{sec:general_sub_gaussian}, we see that the tolerable fraction of adversarial data-points decays fast with $d$, which makes Algorithm~\ref{alg:trimmed_k_means} unsuitable for large $d$. Here we analyze the symmetric $2$-cluster setting only. However given \ref{lem:ini}, our analysis can be extended to general $K$ cluster setting. We adapt a slightly different observation model:  $\{y_i \}_{i=1}^m$ are drawn i.i.d from the following Huber contamination model: with probability $ 1-\alpha $, $y_i = \nu_i (\theta^* + \tau_i)$, where $\nu_i$ is a Rademacher random variable and $\tau_i$ is a $\zeta $-sub-Gaussian random vector with zero mean, and is independent of $\nu_i$; with probability $\alpha $, $y_i$ is drawn from an arbitrary distribution. We assume that the maximum eigenvalue of the covariance matrix of $\tau_i$ is bounded. More specifically, we let $\tilde{\sigma}^2 := \lambda_{\max}(\EE[\tau_i \tau_i^\top])$. We denote the distribution of $y$ and $\tau$ by $\D_y$ and $\D_\tau$, respectively. Intuitively, with probability $1-\alpha $, $y_i$ is an inlier, i.e., drawn from a mixture of two symmetric distributions, and with probability $ \alpha $, $y_i$ is an outlier. The goal is to estimate $\theta^*$ and find the correct labels (i.e., $ \nu_i $) of the inliers. We propose Algorithm~\ref{alg:clustering_iter_filtering} where the total number of data points $m$ is an integer multiple of the number of iterations $T$, and the algorithm uses the Iterative Filtering algorithm~\cite{diakonikolas2016robust,diakonikolas2017being,steinhardt2017resilience}, denoted by $\mathsf{IterFilter}$ as a subroutine. The intuition of the iterative filtering algorithm is to use higher order statistics, such as the sample covariance to iteratively remove outliers.

\begin{algorithm}[h]

  \caption{Clustering with iterative filtering subroutine}
  \begin{algorithmic}[1]
  \REQUIRE Observations $y_1, y_2, \ldots, y_m $, initial guess $\hat{\theta}^{(0)}$, number of iterations $T$.
  \FOR{$s=1,2, \ldots, T$}
  \STATE Label estimation: $\hat{\nu}_i^{(t)} = \mathrm{argmin}_{\nu \in\{-1, 1\}} \norms{\nu y_i - \hat{\theta}^{(t-1)}}^2 $, $ i = \frac{(t-1)m}{T} + 1,., \frac{t m}{T}$.
  \STATE Parameter estimation: $\hat{\theta}^{(t)} = \mathsf{IterFilter}( \{ \hat{\nu}_i^{(t)}y_i\}, i = \frac{(t-1)m}{T} + 1, \frac{(t-1)m}{T} + 2, \ldots, \frac{t m}{T} ) $.
  \ENDFOR
  \end{algorithmic}\label{alg:clustering_iter_filtering}
\end{algorithm}

The convergence guarantee of the algorithm is in Theorem~\ref{thm:him_dim_main}. We start with the following assumption:

\begin{assumption}
\label{asm:high_dim}
We assume: (a) (Initialization) $ \twonms{\hat{\theta}^{(0)} - \theta^*} \le  \frac{1}{2}\norms{\theta^*}$ (b) (SNR) $\norms{\theta^*}/\zeta \ge  C_1$ and (c) (Sample complexity) $ m \ge \frac{C_2}{\alpha}(d + \frac{1}{\alpha}\log(\frac{1}{\eta} \log(\frac{1}{\beta}))\log(\frac{1}{\beta})$.
\end{assumption}
We emphasize that the SNR requirement is standard and the initialization condition is slightly stronger than a random guess. Armed with the above assumption, we have the following result.
\begin{theorem}
\label{thm:him_dim_main}
Suppose that $\alpha \le 1/16 $, $\tilde{\sigma} /\zeta \le C$, and let $\beta := 3\alpha + 8 \exp(-\norms{\theta^*}^2/2\zeta^2)$. With Assumption~\ref{asm:high_dim} and running Algorithm~\ref{alg:clustering_iter_filtering} for $T = \Theta(\log(1/\beta))$ iterations, with probability at least $1-\eta$, we have
\begin{align*}
  \twonms{\hat{\theta}^{(T)} - \theta^*} \le C_3(\tilde{\sigma} + \zeta)(\sqrt{\alpha} + \exp(-\norms{\theta^*}^2/4\zeta^2)),  \,\,\,\,\, \text{and}
\end{align*}
\begin{align*}
\prob{\mathrm{argmin}_{\hat{\nu}\in\{-1, 1\}} \twonms{\hat{\nu}y - \hat{\theta}^{(T)}}^2 \neq \nu \mid y \text{ is inlier}}
  \le C_4(\alpha + \exp(-\norms{\theta^*}^2/2\zeta^2)).
\end{align*}
\end{theorem}
Note that the tolerable level of $\alpha$ has no dependence on dimension, which is an improvement over Theorem~\ref{thm:2cluster}.
\section{Experiments}
\label{sec:simulations}

We perform extensive experiments on synthetic and real data and compare the performance of our algorithm to several non-robust clustering and/or optimization-based algorithms.

\subsection{Synthetic data}
\label{sss:syn_setup}

\begin{figure*}[t]
  \begin{subfigure}[b]{0.32\textwidth}
    \raisebox{5mm}{\centerline{\includegraphics[width=\columnwidth]{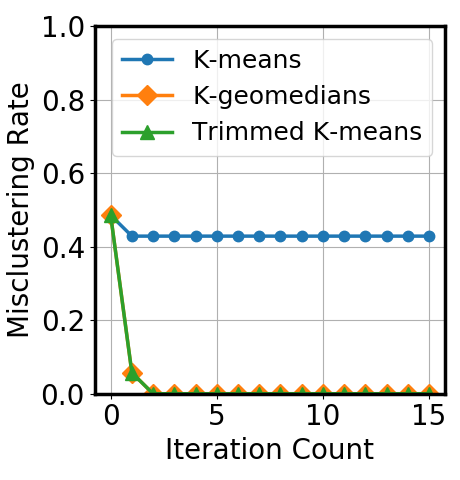}}}
    \caption{Misclustering vs. iteration count}
    \label{fig:perit}
  \end{subfigure}
  \hfill
  \begin{subfigure}[b]{0.32\textwidth}
    \centerline{\includegraphics[width=\columnwidth]{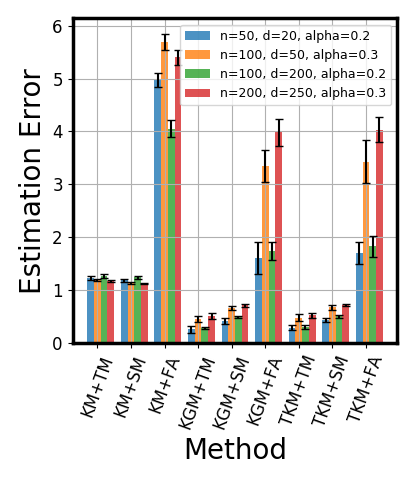}}
    \caption{Synthetic data with $m=100$, \\ $K=5, \sigma=2$}
    \label{fig:synest}
  \end{subfigure}
  \hfill
  \begin{subfigure}[b]{0.32\textwidth}
    \centerline{\includegraphics[width=\columnwidth]{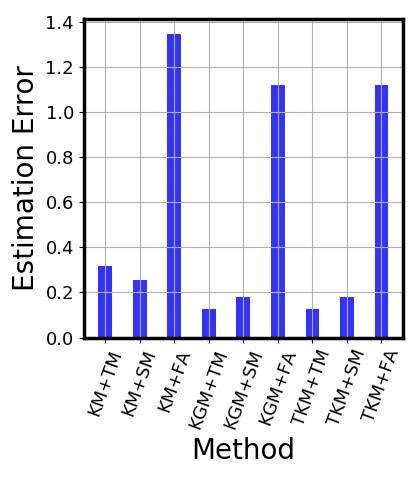}}
    \caption{Yahoo! Learning to Rank \\ Dataset}
    \label{fig:yahooest}
  \end{subfigure}
  \caption{A comparison of K-means (KM), K-geomedians (KGM), and Trimmed K-means (TKM) in conjunction with Trimmed Mean robust optimization (TM), Sample Mean optimization (SM) and Federated Averaging (FA). In Figure~\ref{fig:perit}, we choose $m=100, d=100, K=5, \sigma=3$. The error bars in Figure~\ref{fig:synest} show the standard deviation over 20 trials.}
  \label{fig:synsim}
\end{figure*}

For synthetic experiments, we use a mixture of linear regressions model. For each cluster, a $d$ dimensional regression coefficient vector, $\{\vecw^*_i\}_{i=1}^{K}$, is generated element-wise by a $Bernoulli(\frac{1}{2})$ distribution. Then $\floor{(1-\alpha)m}$ machines are uniformly assigned to the $K$ clusters, and $\lceil\alpha m\rceil$ machines are considered adversarial machines. For each good machine, $j$ (belonging to cluster, $i$), $n$ data points are generated independently according to: $\vecx^{j,l} = \mathbf{\chi}_l^T \vecw_i^* + \tau_l$, for all $l \in [n]$, where $\mathbf{\chi}_l \sim \mathcal{N}(0, I_d)$ and $\tau_l \sim \mathcal{N}(0, \sigma^2)$. For adversarial machines, the regression coefficients are sampled from $3* Bernoulli(\frac{1}{2})$, resulting in outliers. We initialize the cluster assignments with $60$ percent correct assignments for the good machines. We test the performance of Lloyd ($K$-means), Trimmed $K$-means (Algorithm~\ref{alg:trimmed_k_means}) and $K$-geomedians (where the sample mean step of Lloyd is replaced by  geometric median; note that this is Algorithm~\ref{alg:trimmed_k_means} excluding the trimming step). We set $K=5$ and $m=100$. In Figure~\ref{fig:perit}, we see that the fraction of misclustered points (which we call misclustering rate) indeed diminishes with iteration at a fast rate which validates Theorem~\ref{theorem:k_mix}, whereas for $K$-means, it converges to a misclustering rate of $0.4$.

We compare our algorithm consisting of robust clustering (using Trimmed $K$-means or $K$-geomedians) and robust distributed optimization with algorithms without robust subroutines in the clustering or the optimization step. In particular we use the classical $K$-means as a non robust clustering, and a naive sample averaging-based scheme (instead of robust trimmed mean-based scheme by \cite{yin2018byzantine}) as a non-robust, distributed algorithm. Also, in the robust optimization stage, we  compare with a robust version of the Federated Averaging algorithm of \cite{mcmahan2017federated} with $5$ iterations of gradient descent in each worker node before the global model gets updated (by taking the trimmed mean of the local models in the worker nodes).

We first observe that the estimation error ($\max_{i \in [K]} \norms{\hat{\vecw}_i - \vecw^*_i}/\sqrt{d}$) for non-robust clustering schemes (KM in Figure~\ref{fig:synest}) is $\ge 53\%$ higher than that using Trimmed $K$-means (TKM) and $K$-geomedians (KGM). Furthermore, trimmed mean-based distributed optimization (TM) strictly outperforms the sample mean-based (SM) optimization routine by $\ge 29\%$ even with robust clustering. Federated averaging (FA) does orders of magnitude worse in estimation, likely due to the poor gradient updates provided by individual machines. Hence matching our theoretical intuition, robust clustering and robust optimization have the best performance in the presence of adversaries.

\subsection{Yahoo! Learning to Rank dataset}

The performance of the modular algorithm is evaluated on the Yahoo Learning to Rank dataset \cite{yahoo_rank}. We use the \texttt{set2.test.txt} file for our experiment. We choose to treat the data as unsupervised, ignoring the labels for this simulation. Starting with $103174$ queries and $595$ features, we adopt the following thresholding rule: we draw an edge between the queries with $\ell_2$ distance less than $\gamma$ (which we optimize at $3.415$). We then run a tree-search algorithm to detect the connected graphs which produce our true cluster assignments. Small groups are removed from the dataset. This results in $4$ large clusters. Next, we take the mean of the features in each cluster to obtain $\vecw^*_1,\ldots,\vecw^*_4$. The data points in each cluster is then split randomly in batches of $50$ (hence, $n=50$). In addition, respecting $\alpha=0.3$, $80$ adversarial splits are incorporated via sampling $50$ points randomly from the unused data and adding a $Bernoulli(\frac{1}{2}) - 0.5$ vector to the ERM. Note, we synthetically perturb the data points primarily since it is hard to find datasets with explicit adversaries. We then compute the mean in each split (these can be thought analogous to local ERMs), and perform clustering on them using $K$-means, Trimmed $K$-means, and $K$-geomedians algorithms with fully random initialization. Then, we use trimmed mean, sample mean, or Federated Averaging optimization to estimate the $w^*$ on each of the cluster assignment estimates with mean squared loss.

The results of the real data experiments are shown in Fig~\ref{fig:yahooest}. We see that Trimmed $K$-means in conjunction with trimmed mean optimization outperforms the other methods with an estimation error of $0.125$. This algorithm is easy to implement and learns the optimal weights efficiently. On the other hand, the estimation error of $K$-means algorithm with sample mean optimization is $0.256$, which is relatively two times worse than the robust algorithms. Also, Trimmed $K$-means and $K$-geomedians have similar final estimation error, which further confirms that trimming step after computing the geometric median may be redundant. Thus, we once again emphasize that our robust algorithm performs better than standard non-robust algorithms.

\section{Conclusion and future work}
We tackle the problem of robust Federated Learning in a heterogeneous environment. We propose a $3$-step modular solution to the problem. For the second step, we analyze the classical Lloyd algorithm with a robust subroutine. Weakening the sub-Gaussian assumption along with a better initialization scheme are kept as our future endeavors.

\section*{Acknowledgments}
The authors would like to thank Swanand Kadhe and Prof. Peter Bartlett for helpful discussions.

\bibliographystyle{unsrt}
\bibliography{ref}

\appendix

\section*{Appendix}
 
 \section{Theoretical guarantees for Algorithm \ref{alg:trimmed_k_means}}
\label{appendix:robust_clustering}

\subsection{Proof of Proposition~\ref{prop:sub_Gaussian}}
Given the parametric form of data generation, we first stack the covariates to form the matrix $Z \in \R^{n \times d} $ where $Z^\top := [ \mathbf{\chi}_1^\top \,\, \mathbf{\chi}_2^\top \ldots \mathbf{\chi}_n^\top]$. Also we form the vectors $X_j^\top := [\vecx^{i,1} \,\, \vecx^{i,2} \ldots \vecx^{i,n}]$ and $\Upsilon^T :=[\Upsilon_1 \ldots \Upsilon_n]$. The objective is to estimate $\vecw^*_k$. We run an ordinary least squares, i.e., we calculate the following,
\begin{align*}
\hat{\vecw}^{(i)} = \arg \min_\vecw \norms{X_j - Z \vecw}^2
\end{align*} 
From standard calculations, The ERM $\hat{\vecw}^{(i)}$ is given by
\begin{align*}
\hat{\vecw}^{(i)} = (Z^\top Z)^{-1} Z^\top (Z \vecw^*_k + \Upsilon) = \vecw^*_k + (Z^\top Z)^{-1} Z^\top \Upsilon
\end{align*}
Hence the distribution of $\hat{\vecw}^{(i)}$ is Gaussian. Since $\mathbb{E}(\Upsilon_j)=0$ for all $j \in [n]$, $\mathbb{E}(\hat{\vecw}^{(i)})= \vecw^*_k$.

\subsection{Symmetric $2$ cluster: proof of Theorem~\ref{thm:2cluster}}

Suppose after geometric median based trimming on both the centers at iteration $s$, we retain $(1-\beta)m$  data points.

Let $\hat{\theta}^{(s)}$ be the estimate of $\theta^*$ at iteration $s$. Let us fix a few notations here. At step $s$, we denote $\mathcal{U}$ as the set of data-points that are not trimmed. $\mathcal{T}$ denotes the set of trimmed points and $\mathcal{B}$ denotes the set of adversarily corrupted data points. We have

\begin{eqnarray}
\hat{\theta}^{(s)} & = & \frac{1}{(1-\beta)m}\sum_{i \in \mathcal{U}} z_i^{(s)}y_i \nonumber \\
& = & \frac{1}{(1-\beta)m} \bigg [ \sum_{i \in \mathcal{M}} z_i^{(s)}y_i - \sum_{i \in \mathcal{M} \cap \mathcal{T}} z_i^{(s)}y_i + \sum_{i \in \mathcal{B} \cap \mathcal{U}} z_i^{(s)}y_i \bigg ] \nonumber \\
& = & \frac{1}{(1-\beta)m} \sum_{i \in \mathcal{M}} z_i^{(s)}y_i  - T_1 + T_2
\end{eqnarray}

where, $T_1 =  \frac{1}{(1-\beta)m}\sum_{i \in \mathcal{M} \cap \mathcal{T}} z_i^{(s)}y_i $ and $T_2 = \frac{1}{(1-\beta)m} \sum_{i \in \mathcal{B} \cap \mathcal{U}} z_i^{(s)}y_i $. Consequently

\begin{eqnarray}
\hat{\theta}^{(s)} - \theta^* & = & \frac{1}{(1-\beta)m} \sum_{i \in \mathcal{M}} z_i^{(s)}y_i  - T_1 + T_2 -\theta^* \nonumber \\
& = & \frac{1}{(1-\beta)m} \sum_{i \in \mathcal{M}} z_i^{(s)}y_i -\frac{1}{(1-\beta)m} \sum_{i \in \mathcal{M}} z_i y_i +\frac{1}{(1-\beta)m} \sum_{i \in \mathcal{M}} z_i y_i -T_1 +T_2 -\theta^*. \nonumber
\end{eqnarray}

Since $z_iy_i = \theta^* +\xi_i$, where $z_i$ are the true label of the $i$-th data point, we have the following relation
$$
z_i^{(s)}-z_i = -2 I\{z_i^{(s)} \neq z_i \} z_i
$$

Let $\gamma:=\frac{1-\alpha}{1-\beta}$. Plugging in, we get
 \begin{eqnarray}
&& \hat{\theta}^{(s)} - \theta^*  =  \frac{1}{(1-\beta)m} \sum_{i \in \mathcal{M}} -2 I\{z_i^{(s)} \neq z_i \}(\theta^* +\xi_i)+\frac{1}{(1-\beta)m} \sum_{i \in \mathcal{M}} z_i y_i -T_1 +T_2 -\theta^* \nonumber \\
& = & \frac{1}{(1-\beta)m} \bigg [ \sum_{i \in \mathcal{M}} -2 I\{z_i^{(s)} \neq z_i \}  (\theta^* +\xi_i)+\sum_{i \in \mathcal{M}} (\theta^*+\xi_i) \bigg ] - T_1+T_2 -  \theta^*   \nonumber \\
& = & \gamma \frac{1}{(1-\alpha)m} \bigg [ \sum_{i \in \mathcal{M}} -2 I\{z_i^{(s)} \neq z_i \} (\theta^* +\xi_i) + \sum_{i \in \mathcal{M}} \xi_i \bigg ] -T_1 +T_2 + (\gamma -1) \theta^*.
 \end{eqnarray}

We need a few definitions to proceed further. Recall $A_s = \frac{1}{(1-\alpha)m} \sum_{i \in \mathcal{M}} I\{z_i^{(s)} \neq z_i \} $ denote the average error rate over good samples. Also define $R:= \frac{1}{(1-\alpha)m} \sum_{i \in \mathcal{M}}  \xi_i$ and $\bar{\tau}:=\frac{1}{(1-\alpha)m} \sum_{i \in \mathcal{M}} \xi_i$. With the above definitions, we get
$$
\hat{\theta}^{(s)} - \theta^* = \gamma (-2A_s \theta^* -2R  + \bar{\tau}) -T_1 +T_2 + (\gamma -1) \theta^*
$$

As a result
\begin{eqnarray}
\langle \hat{\theta}^{(s)}, \theta^* + \xi_i \rangle = \gamma \langle \theta^* + \xi_i, (1-2A_s) \theta^* -2R +\bar{\tau} \rangle  - \langle \theta^* +\xi_i, T_1-T_2 \rangle \nonumber
\end{eqnarray}

The extra term $\langle \theta^* +\xi_i, T_1-T_2 \rangle$ can be controlled using Cauchy Schwartz inequality in the following way
\begin{eqnarray}
\langle \theta^* +\xi_i, T_1-T_2 \rangle \leq \norms{\theta^* +\xi_i} \norms{T_1-T_2} \leq (\norms{\theta^*}+\norms{\xi_i})(\norms{T_1}+\norms{T_2}) \nonumber
\end{eqnarray}
where the last inequality follows from triangle inequality. Furthermore
\begin{align*}
\norms{T_2} \leq \frac{\alpha}{1-\beta} \max_{i \in \mathcal{\mathcal{U}}} \norms{y_i} \,\,\, \text{and}
\end{align*}
\begin{align*}
\norms{T_1 } \leq \frac{\beta}{1-\beta}\max_{i \in \mathcal{M}}\norms{y_i}.
\end{align*}

As a result
\begin{align}
\label{eqn:norm_extra_term}
\norms{T_2} + \norms{T_1} \leq \frac{1}{1-\beta} \left( \beta \max_{i \in \mathcal{M}}\norms{y_i} + \alpha \max_{i \in \mathcal{U}}\norms{y_i} \right)
\end{align}

Let us first control the second term of the above equation. Let $\mu_{geo}$ denote the geometric median of the data points. From Algorithm~\ref{alg:trimmed_k_means}, for all $i \in \mathcal{U}$, we have
\begin{align*}
\norms{y_i - \mu_{geo}} \leq C \sigma \sqrt{d} \Rightarrow \norms{y_i - \theta^* + \theta^* -\mu_{geo}} \leq C \sigma \sqrt{d}
\end{align*}
Invoking \cite[Theorem 3.1]{minsker2015}, we obtain
\begin{align*}
\norms{\mu_{geo}-\theta^*} \leq C_1 \sigma \sqrt{d}
\end{align*}
with probability at least $1-c \exp(-m)$. Now, using a triangle inequality, we obtain
\begin{align*}
\norms{y_i - \theta^*} \leq (C+C_1) \sigma \sqrt{d}
\end{align*}
which upon further modification yields
\begin{align*}
\norms{y_i} \leq \norms{\theta^*} + (C+C_1) \sigma \sqrt{d}
\end{align*}
For the first term of Equation~\ref{eqn:norm_extra_term}, we just substitute $y_i=z_i\theta^* + w_i$ to obtain
\begin{align*}
\max_{i \in \mathcal{M}}\norms{y_i} = \norms{\theta^*}+\max_{i \in \mathcal{M}} \norms{\xi_i}
\end{align*}
Now $\xi_i$ is a gaussian random vector in dimension $d$ with i.i.d gaussian coordinates. Therefore, the distribution of squared norm $\norms{\xi_i}^2 \sim \chi_d^2$ (a Chi-squared random variable of degree $d$). Also, since $\norms{\xi_i} \geq 0$, the term $\max_{ i \in \mathcal{M}}\norms{\xi_i}^2 = (\max_{i \in \mathcal{M}} \norms{\xi_i})^2$. From $\chi_d^2$ concentration (\cite{boucheron}) and taking a union bound,  for $t \in (0,1)$, we get
\begin{eqnarray}
\mathbb{P}\{ \max_{i \in \mathcal{M}} \norms{\xi_i}^2 \geq d(\sigma^2 +t) \} \leq (1-\alpha)m \exp(-dt^2) \nonumber
\end{eqnarray}

Therefore with probability at least $1- (1-\alpha)m \exp(-dt^2)$, the quantity $\max_{i \in \mathcal{M}}\norms{\xi_i} \leq \sqrt{d}(\sigma + \sqrt{t})$. From the same $\chi_d^2$ concentration, we get $\norms{\xi_i} \leq \sqrt{d}(\sigma + \sqrt{t_1})$ with probability at least $1-\exp(-d t_1^2)$ for $t_1 \in (0,1)$. Substituting, and using $\alpha \leq \beta$, we obtain
\begin{eqnarray}
\langle \theta^* +\xi_i, T_1-T_2 \rangle & \leq & \frac{2\beta}{1-\beta} \bigg ( \norms{\theta^*}^2 + C_1 \sqrt{d} \sigma \norms{\theta^*} + C_2 \sigma^2 d \bigg ) \nonumber \\
& \leq & \frac{2\beta}{1-\beta} \norms{\theta^*}^2 \bigg ( 1+ \frac{C_1 \sigma \sqrt{d}}{\norms{\theta^*}}+ \frac{C_2 \sigma^2 d}{\norms{\theta^*}^2} \bigg ). \nonumber
\end{eqnarray}
Since $\alpha \leq \beta \leq \frac{c}{d}$, we have
\begin{align*}
\langle \theta^* +\xi_i, T_1-T_2 \rangle \leq \norms{\theta^*}^2 \left(\frac{2c}{d(1-c/d)}+ \frac{2c C_1}{r \sqrt{d}(1-c/d)}+ \frac{2c C_2}{r^2(1-c/d)} \right)
\end{align*}
 Define $\Gamma:=\left(\frac{2c}{d(1-c/d)}+ \frac{2c C_1}{r \sqrt{d}(1-c/d)}+ \frac{2c C_2}{r^2(1-c/d)} \right)$. We have,
\begin{align*}
\langle \theta^* +\xi_i, T_2-T_1 \rangle \leq \Gamma \norms{\theta^*}^2
\end{align*}
With Assumption~\ref{asm:snr}, we get
\begin{align*}
\Gamma \leq \left(\frac{2c}{d(1-c/d)}+ \frac{2c C_1}{C_{th} \sqrt{d}(1-c/d)}+ \frac{2c C_2}{C_{th}^2 (1-c/d)} \right)
\end{align*}

 Also, from Lemma~\ref{lem:gauss_7.1} applied to the set $S=\{ i \in [(1-\alpha)m]:  I\{z_i^{(s+1)} \neq z_i^{(s)} \}=1 \}$, we get $\norms{R} \leq \frac{\norms{\theta^*}}{r}\sqrt{2A_s}$. Invoking Lemma~\ref{lem:gauss_7.2}, we have
$$
\langle 2R-\bar{\tau},\theta^* \rangle \leq \frac{\norms{\theta^*}^2}{r}\sqrt{2A_s} + \frac{\norms{\theta^*}^2}{\sqrt{(1-\alpha)m}}
$$

 Now, we analyze the error in the $s+1$-th step. The iteration is still the nearest neighbor assignment. Hence

\begin{eqnarray}
  z_i^{(s+1)} = \underset{r \in \{ -1,+1 \} }{\mathrm{argmin}}\norms{ry_i - \hat{\theta}^{(s)}}^2=\underset{r \in \{-1,+1\}}{\mathrm{argmax}} \langle ry_i,\hat{\theta}^{(s)} \rangle =\underset{r \in \{-1,+1 \}}{\mathrm{argmax}} \langle\theta^*+\xi_i,\hat{\theta}^{(s)} \rangle .\nonumber
\end{eqnarray}

From this, the term $I\{z_i^{(s+1)} \neq z_i^{(s)} \} = I\{ \langle \theta^*+\xi_i,\hat{\theta}^{(s)} \rangle \leq 0 \}$. Using the above calculation
\begin{eqnarray}
I\{z_i^{(s+1)} \neq z_i^{(s)} \} \leq I \{\gamma \bigg ( \Omega_0 \norms{\theta^*}^2 + \langle \xi_i,\theta^* \rangle + \langle \xi_i,-2A_s \theta^*+2R-\bar{\tau} \rangle \bigg) \leq 0 \}. \label{eqn:mistake_indicator}
\end{eqnarray}

where $\Omega_0 = 1-2A_s - \frac{2\sqrt{2A_s}}{r}-\frac{1}{\sqrt{(1-\alpha)m}}-\frac{\Gamma}{\gamma}$. A naive upper bound on $\Omega_0$ is the following
$$
\beta_0 \geq 1-2A_s-\frac{2}{r}-\frac{1}{\sqrt{(1-\alpha)m}}-\frac{\Gamma}{\gamma}.
$$

From the expression of $\Gamma$ and using $\beta \geq \alpha$, we obtain
\begin{align*}
\frac{\Gamma}{\gamma} \leq \left(\frac{2c}{d(1-c/d)}+ \frac{2c C_1}{C_{th} \sqrt{d}(1-c/d)}+ \frac{2c C_2}{C_{th}^2 (1-c/d)} \right).
\end{align*}

 Also, we choose $C_{th}$ sufficiently large to ensure that $\frac{\Gamma}{\gamma} \leq \varepsilon$ with high probability, where $\varepsilon$ is a (small) positive constant.

Since $\gamma > 0$, we can drop it inside the indicator function of Equation~\ref{eqn:mistake_indicator}. Now we are ready to prove the theorem.

\subsubsection{Proof of the first part}

 For $a,b \in \R$ and $c > 0$, we use the following inequality on the indicator function
 $$ I\{a+b \leq 0\} \leq I\{a\leq c\}+I\{b \leq -c\} \leq I\{a \leq c\}+\frac{b^2}{c^2}.
 $$
Using this, we have
\begin{eqnarray}
I\{z_i^{(s+1)} \neq z_i^{(s)} \} \leq I\{\Omega \norms{\theta^*}^2 \leq -\langle \xi_i,\theta^* \rangle \} +\frac{\langle \xi_i,2R-\bar{\tau}-2A_s\theta^* \rangle^2}{\delta^2\norms{\theta^*}^4} \nonumber
\end{eqnarray}

where we define $\Omega = \Omega_0 - \delta$ with $\delta = \frac{3.12}{r}$. We now take the average over all good data points and obtain $A_s \leq I_1 + I_2$ where

\begin{equation}
I_1 = \frac{1}{(1-\alpha)m}\sum_{i=1}^{(1-\alpha)m} I\{\langle \xi_i,\theta^* \rangle \leq -\Omega \norms{\theta^*}^2 \}
\end{equation}

\begin{equation}
I_2 =  \frac{1}{(1-\alpha)m\delta^2 \norms{\theta^*}^4} \sum_{i=1}^{(1-\alpha)m} \langle \xi_i,2R-\bar{\tau}-2A_s\theta^* \rangle^2
\end{equation}

\textit{Upper Bound on $I_1$}:

We define $\eta_a = 1-2a -\frac{5.12}{r}-\frac{1}{\sqrt{(1-\alpha)m}}-\frac{\Gamma}{\gamma}$, and $a \in \mathcal{D}$, where $\mathcal{D}$ is the set of discrete values $A_s$ can take. We have
$$
\mathcal{D}= \left \{ \frac{1}{(1-\alpha)m}, \frac{2}{(1-\alpha)m}, \ldots, \frac{\floor{(1-\alpha)m/2}}{(1-\alpha)m } \right \}.
$$

With this, we observe that $I\{ \langle \xi_i, \theta^* \rangle \leq -\eta_a \norms{\theta^*}^2 \}$ for all $i \in [(1-\alpha)m]$ are independent Bernoulli random variable. Using Hoeffding's inequality (see e.g. Theorem 3.2 of \cite{lu2016statistical}), we get the following upper bound on $I_1$
\begin{equation}
I_1 \leq \exp \bigg ( -\frac{\eta_{A_s}^2\norms{\theta^*}^2}{2\sigma^2} \bigg ) + \sqrt{\frac{4\log((1-\alpha)m/2)}{(1-\alpha)m}}
\end{equation}
with probability at least $1-\frac{1}{((1-\alpha)m)^3}$.
\vspace{3mm}

\textit{Upper Bound on $I_2$}:
We replace the parameter $m$ by $m':=(1-\alpha)m$, since this is the effective sample size. We follow \cite{lu2016statistical} (Theorem 3.2) and use Lemma~\ref{lem:gauss_7.3} with $m'$. We get the following bound
\begin{equation}
I_2 \leq \frac{8}{r^2}A_s + \frac{1}{r^2} + \frac{1}{(1-\alpha)m}+A_s^2
\end{equation}
with probability exceeding $1-2(1-\alpha)m\exp(-C_1 d) - \exp(-C_2d) -2\exp(-\frac{\norms{\theta^*}^2}{3\sigma^2})$.

Combining all the pieces
$$
A_{s+1} \leq \exp \bigg ( -\frac{\eta_{A_s}^2\norms{\theta^*}^2}{2\sigma^2} \bigg ) + \sqrt{\frac{4\log((1-\alpha)m/2)}{(1-\alpha)m}}+ \frac{8}{r^2}A_s + \frac{1}{r^2} + \frac{1}{(1-\alpha)m}+A_s^2
$$
Let $p=\frac{1}{2}-\frac{2.56}{r}-\frac{1}{2\sqrt{(1-\alpha)m}}-\frac{\varepsilon}{2}$. From above, if $A_0 \leq p $, and if the constants $C_{th}'$ and $m_{th}$ are sufficiently large, we show via induction argument that $A_s \leq p$ for all $s$ and furthermore, we have $ \eta_{A_s} \geq \frac{2\sqrt{\log r}}{r} $
for all $s \geq 0$. Plugging in, we get

\begin{equation}
A_{s+1} \leq A_s (A_s + \frac{8}{r^2}) + \frac{2}{r^2} +  \sqrt{\frac{4\log((1-\alpha)m)}{(1-\alpha)m}} \nonumber
\end{equation}
with probability at least $1-\frac{1}{(m(1-\alpha))^3}-2(1-\alpha)m\exp(-C_1 d) - \exp(-C_2 d) -2\exp(-\frac{\norms{\theta^*}^2}{3\sigma^2})$.

\subsubsection{Proof of the second part}

By the above bound on $A_{s+1}$, if $C_{th}$ is sufficiently large,
$$
A_{s+1} \leq \frac{1}{2}A_s + \frac{2}{r^2}+\sqrt{\frac{4\log m}{m}}
$$
Iterating, we get, $A_s \leq \frac{4}{r^2} + 5\sqrt{\frac{\log m}{m}}$ for all $s \geq \log m$. We now analyze the mistake bound $I\{z_i^{(s+1)} \neq z_i^{(s)}\}$ for $s \geq \log m $ similar to the proof in Section 7.3 of \cite{lu2016statistical}, which yields the following. For a sufficiently large $m_{th}$, with probability at least $1-\frac{C_3}{m(1-\alpha)}-2(1-\alpha)m\exp(-C_1 d) - \exp(-C_2d) -C_4\exp(-\frac{\norms{\theta^*}^2}{16\sigma^2})$
$$
A_s \leq \exp \bigg ( -\frac{\norms{\theta^*}^2}{16\sigma^2} \bigg )
$$

\subsection{Details of $K$-clusters with sub-Gaussian mixture}
\label{sec:details_gen_K}
 We introduce a few new notations; let $\Delta = \min_{g\neq h \in [K]} \norm{\theta_g - \theta_h}$ be the signal strength, i.e.\  the minimum separation between clusters. For cluster $h$, let $\hat{\theta}_h^{(s)}$ be the estimate of $\theta_h$ at iteration $s$. Define $\lambda = \max_{g\neq h\in [K]} \norm{\theta_g - \theta_h}/\Delta$, the maximum signal strength relative to the minimum. Our error rate of the center estimates can be measured by $ \Lambda_s = \max_{h\in [K]} \frac{1}{\Delta}\norm{\hat{\theta}_h^{(s)} - \theta_h}$.

 Let $T^*_h$ and $T^{(s)}_h$ be the set of nodes in the true cluster $h$ and the estimated cluster $h$ at step $s$, respectively. Now, we define $S_{gh}^{(s)} = T_g^* \cap T_h^{(s)}$, the set of nodes in cluster $g$ estimated to be in cluster $h$ at step $s$. We define the cardinality of these sets as $m_h^* = |T_h^*|$, $m_h^{(s)} = |T_h^{(s)}|$, $m_{gh}^{(s)} = |S_{gh}^{(s)}|$. For simplicity, we will omit the superscript $(s)$ when working at a single step of the algorithm. Let $\mathcal{B}$ represent the set of adversarial nodes, and recall that $\alpha$ is the fraction of adversarial nodes, i.e. $\alpha = \frac{|\mathcal{B}|}{m}$. For cluster $h$, let $\alpha_h^{(s)}$ be the fraction of adversarials in $T_h^{(s)}$. We define a cluster-wise mis-clustering fraction at iteration s as
 \begin{align}
 \label{eqn:mis_cluster_rate}
 G_s = \max_{h\in [K]}\left\{ \frac{\sum_{g\neq h\in [K]}m_{gh}^{(s)}}{(1-\alpha_h)m_h^{(s)}}, \frac{\sum_{g\neq h\in [K]}m_{hg}^{(s)}}{m_h^*}\right\}.
\end{align}
 Let the set of untrimmed nodes at a given time step be $\mathcal{U}$, and the set of trimmed nodes as $\mathcal{T}$. For cluster $h$, $\mathcal{U}_h, \mathcal{T}_h$ represent the untrimmed and trimmed sets relative to the geomedian of $\hat{\theta}_h$ respectively. Now, let the superscript $\mathcal{U}$ represent the set contained in the untrimmed set for a given cluster. For example, $T_{h}^{\mathcal{U}} = T_h \cap \mathcal{U}_h$, and $S_{gh}^{\mathcal{U}} = T_h^{\mathcal{U}} \cap T_g^*$. Similarly, $m_h^{\mathcal{U}} = |T_h^{\mathcal{U}}|$ and $m_{gh}^{\mathcal{U}} = |T_g^* \cap T_h^{\mathcal{U}}|$.

 Let $\beta$ be the fraction of nodes trimmed by the algorithm at a certain step, i.e. $\beta = \frac{|\mathcal{T}|}{m}$. For cluster $h$, let $\beta_h$ be the fraction trimmed from the ball centered at $\hat{\theta}_h$. In addition, to express the adversarial fraction within an untrimmed set, let $\alpha_h^{\U(s)}$ be the fraction of adversarials in $T_h^{\U(s)}$. We also define $\alpha' := \max_h \alpha_h^{\U(s)}$. We use $\alpha'$ to in the initialization as well as to upper bound the fraction of mis clustered points.

With respect to our trimmed-mean algorithm, we define a modified trimmed cluster-wise mis-clustering rate at iteration $s$ as
\begin{align*}
G_s^{\mathcal{U}} = \max_{h\in [K]}\left\{ \frac{\sum_{g\neq h\in [K]}m_{gh}^{\mathcal{U}(s)}}{(1-\alpha_h^\U)m_h^{\mathcal{U}(s)}}, \frac{\sum_{g\neq h\in [K]}m_{hg}^{(s)} + m_{hh}^{\mathcal{T}(s)}}{m_h^*}\right\}
\end{align*}
Define the minimum cluster size as $\gamma_1 = \min_{h\in [K]}\frac{m_h^*}{(1-\alpha)m}$. Lastly, we define a normalized signal-to-noise ratio for $K$ clusters as $ r_1 = \frac{\Delta}{\sigma}\sqrt{\frac{\gamma_1}{1 + \frac{Kd}{(1-\alpha)m}}}$.

With the above notations, invoking Assumption~\ref{asm:k_mix} and Theorem~\ref{theorem:k_mix} ensures an exponential decay of $G_S$.

\subsection{General $K$ cluster: proof of Theorem~\ref{theorem:k_mix}}

We begin by analyzing the following two results which will be crucial to prove the theorem. Let $\mathcal{E}$ be the event of the intersection of Lemma~\ref{lem:sub_gauss_A.1}, \ref{lem:sub_gauss_A.2}, \ref{lem:sub_gauss_A.4} and \ref{lem:sub_gauss_A.5}. We have,

\begin{lemma}
\label{lemma:a6}
  On event $\mathcal{E}$, if $G_s^\mathcal{U} \le \frac{1}{2}$, then we have
  \begin{align*}
    \Lambda_s \le \frac{3}{r_1} + \frac{2\alpha' C\sigma \sqrt{d}}{\Delta} + \min\left\{\frac{3}{r_1}\sqrt{kG_s^\mathcal{U}} + 2G_s^\mathcal{U}\Lambda_{s-1}, \lambda G_s^\mathcal{U}\right\}
  \end{align*}
  where $\alpha'^{(s)} = \max_{h\in [k]} \alpha_h^{\U(s)}$.
\end{lemma}

Next, we present a bound on the misclustering rate per iteration.

\begin{lemma}
\label{lemma:a7}
  On event $\mathcal{E}$, if $\Lambda_s \leq \frac{1-\epsilon}{2}\:and\: r_1 \ge 36\epsilon^{-2}$, then
  \begin{align*}
    G_{s+1}^\U \leq \frac{2}{\epsilon^4r_1^2} + \left(\frac{28}{\epsilon^2r_1}\Lambda_s\right)^2 + \sqrt{\frac{5k\log((1-\alpha)m)}{\gamma_1^2(1-\alpha)m}}
  \end{align*}
\end{lemma}

\subsubsection{Proof of Lemma~\ref{lemma:a6}}

  Let $\bar{Y}_B = \frac{1}{|B|}\sum_{i\in B} y_i$. We begin by expanding the error of estimated centers at timestep $s$. For some cluster $h$:
\begin{align*}
  \hat{\theta}_h - \theta_h &= \frac{1}{m_h^\U}\left(\sum_{i\in S_{hh}^\U} (y_i - \theta_h) + \sum_{a \neq h} \sum_{i\in S_{ah}^\U} (y_i - \theta_h) + \sum_{i\in S_{\mathcal{B}h}^\U} (y_i - \theta_h)\right) \\
                            &= \frac{1}{m_h^\U}\sum_{i\in S_{hh}^\U} \tau_i + \sum_{a\neq h} \frac{m_{ah}^\U}{m_h^\U}(\bar{Y}_{S_{ah}^\U} - \theta_h) + \frac{1}{m_h^\U} \sum_{i\in S_{\mathcal{B}h}^\U}(y_i - \theta_h) \\
\end{align*}
By the label update step of the algorithm, we know $\norm{y_i - \hat{\theta}_h^{(s-1)}} \leq \norm{y_i - \hat{\theta}_a^{(s-1)}}$ for any $i \in S_{ah}$ and in turn $i \in S_{ah}^\U$. Taking the average we have
\begin{align*}
  \norm{\bar{Y}_{S_{ah}^\U} - \hat{\theta}_h^{(s-1)}} \leq \norm{\bar{Y}_{S_{ah}^\U} - \hat{\theta}_a^{(s-1)}}
\end{align*}
By the triangle inequality
\begin{align*}
  \norm{\bar{Y}_{S_{ah}^\U} - \theta_h} \leq \norm{\bar{Y}_{S_{ah}^\U} - \theta_a} + \norm{\hat{\theta}_a^{(s-1)} - \theta_a} + \norm{\hat{\theta}_h^{(s-1)} - \theta_h}
\end{align*}
Let $W_B = \sum_{i \in B} \tau_i$. By Lemma~\ref{lem:sub_gauss_A.1} and substituting in the definition of $\Lambda_{s-1}$
\begin{align*}
  \norm{\bar{Y}_{S_{ah}^\U} - \theta_h} &\leq \norm{W_{S_{ah}^\U}} + 2\Lambda_{s-1}\Delta \\
                                        &\leq \frac{\sigma \sqrt{3((1-\alpha)m + d)}}{\sqrt{m_{ah}^\U}} + 2\Lambda_{s-1}\Delta
\end{align*}
We take a weighted sum over $a\neq h \in [k]$ to get
\begin{align*}
  \sum_{a\neq h} \frac{m_{ah}^\U}{m_h^\U}\norm{\bar{Y}_{S_{ah}^\U} - \theta_h} &\leq \frac{\sigma \sqrt{3((1-\alpha)m + d)}}{m_h^\U}\sum_{a\neq h}\sqrt{m_{ah}^\U} + 2\Lambda_{s-1}\Delta \sum_{a\neq h} \frac{m_{ah}^\U}{m_h^\U} \\
                                                                    &\leq \frac{\sigma \sqrt{3((1-\alpha)m + d)}}{\sqrt{m_h^\U}}\sqrt{(k-1)G_s^\U} + 2G_s^\U\Lambda_{s-1}\Delta
\end{align*}
To bound $\norm{W_{S_{hh}^\U}}$ we use the fact that $W_{S_{hh}^\U} = W_{T_h^*} - \sum_{a\neq h} W_{S_{ha}} - W_{S_{hh}^\mathcal{T}}$. By the triangle inequality and Lemma~\ref{lem:sub_gauss_A.1} and \ref{lem:sub_gauss_A.4}, we have
\begin{align*}
  \norm{W_{S_{hh}^\U}} &\leq \norm{W_{T_h^*}} + \norm{\sum_{a\neq h} W_{S_{ha}} + W_{S_{hh}^\mathcal{T}}} \\
                       &\leq 3\sigma \sqrt{(d + \log (1-\alpha)m)m_h^*} + \sigma \sqrt{3((1-\alpha)m + d)(m_h^* - m_{hh}^\U)}
\end{align*}
By our trimmed mean algorithm, any node within the untrimmed ball is within $C\sigma \sqrt{d}$ from the geometric median. Thus, in the worst case, the adversarial fraction will increase the error by $2C\sigma \sqrt{d}$ each. So we can bound the error in the adversarial fraction like so
\begin{align*}
  \frac{1}{m_h^\U} \norm{\sum_{i \in S_{\mathcal{B}h}^\U}(y_i - \theta_h)} \leq 2\alpha_h^\U C\sigma \sqrt{d}
\end{align*}
With the condition $G_s^\U \leq \frac{1}{2}$, we can lower bound $m_h^\U$.
\begin{align*}
  m_h^\U \ge m_{hh}^\U \ge m_h^*(1-G_s^\U) \ge \frac{1}{2}m_h^* \ge \frac{1}{2}\gamma_1 (1-\alpha)m
\end{align*}
Putting it all together, we have
\begin{align*}
  \begin{split}
    \norm{\hat{\theta}_h^{(s)} - \theta_h} \leq &\frac{3\sigma \sqrt{(d + \log (1-\alpha)m)m_h^*}}{m_h^\U} + \frac{\sigma \sqrt{3((1-\alpha)m + d)(m_h^* - m_{hh}^\U)}}{m_h^\U}  \\
                                           & + \frac{\sigma \sqrt{3((1-\alpha)m + d)m_h^\U(k-1)G_s^\U}}{m_h^\U} + 2G_s^\U\Lambda_{s-1}\Delta + 2\alpha_h^\U C\sigma \sqrt{d}
  \end{split} \\
  \leq & \frac{3\sigma \sqrt{d + \log (1-\alpha)m}}{\sqrt{\gamma_1 (1-\alpha)m}} + \frac{3\sigma \sqrt{((1-\alpha)m + d)(m_h^*-m_{hh}^\U+ m_h^\U(k-1)G_s^\U)}}{m_h^\U} \\
  & + 2G_s^\U\Lambda_{s-1}\Delta + 2\alpha_h^\U C\sigma \sqrt{d} \\
  \leq & \frac{3\sigma \sqrt{d + \log (1-\alpha)m}}{\sqrt{\gamma_1 (1-\alpha)m}} + \frac{3\sigma \sqrt{k((1-\alpha)m + d)G_s^\U}}{\sqrt{\gamma_1 (1-\alpha)m}} \\
  & + 2G_s^\U\Lambda_{s-1}\Delta + 2\alpha_h^\U C\sigma \sqrt{d} \\
  \leq & \left(\frac{3}{r_1}(1 + \sqrt{kG_s^\U}) + 2G_s^\U\Lambda_{s-1}\right) \Delta + 2\alpha_h^\U C \sigma \sqrt{d}
\end{align*}
This gives us the first term in the right hand side (RHS) of the lemma. For the second term we start with a different decomposition of the center estimate.
\begin{align*}
  \hat{\theta}_h &= \frac{1}{m_h^\U}\sum_{i=1}^m (\theta_{z_i} + \tau_i)\mathbbm{1}{\{\hat{z}_i = h \cap i \in \U_h\}} + \frac{1}{m_h^\U} \sum_{i\in S_{\mathcal{B}h}} y_i \\
                 &= \frac{1}{m_h^\U}\sum_{a=1}^k \sum_{i=1}^m \theta_a \mathbbm{1}{\{z_i = a \cap \hat{z}_i = h \cap i \in \U_h\}} + \frac{1}{m_h^\U}W_{T_h^\U} + \frac{1}{m_h^\U} \sum_{i\in S_{\mathcal{B}h}} y_i \\
                 &= \sum_{a=1}^k \frac{m_{ah}^\U}{m_h^\U} \theta_a + \frac{1}{m_h^\U}W_{T_h^\U} + \frac{1}{m_h^\U} \sum_{i\in S_{\mathcal{B}h}} y_i
\end{align*}
We use this bound the error of the estimate.
\begin{align*}
  \norm{\hat{\theta}_h^{(s)} - \theta_h} &= \norm{\sum_{a=1}^k \frac{m_{ah}^\U}{m_h^\U} (\theta_a - \theta_h) + \frac{1}{m_h^\U}W_{T_h^\U} + \frac{1}{m_h^\U} \sum_{i\in S_{\mathcal{B}h}} (y_i - \theta_h)} \\
                                         &\leq \norm{\sum_{a\neq h} \frac{m_{ah}^\U}{m_h^\U}(\theta_a - \theta_h)} + \norm{\frac{1}{m_h}W_{T_h^\U}} + \norm{\frac{1}{m_h^\U}\sum_{i\in S_{\mathcal{B}h}} (y_i - \theta_h)}
\end{align*}
By the triangle inequality we can bound the first term as
\begin{align*}
  \norm{\sum_{a\neq h} \frac{m_{ah}^\U}{m_h^\U}(\theta_a - \theta_h)} &\leq \sum_{a\neq h} \frac{m_{ah}^\U}{m_h^\U}\norm{\theta_a - \theta_h} \\
                                                                      &\leq \lambda\Delta\sum_{a\neq h} \frac{m_{ah}^\U}{m_h^\U} \\                                                                      &\leq \lambda\Delta G_s^\U
\end{align*}
Using Lemma~\ref{lem:sub_gauss_A.1} and the above, we get
\begin{align*}
  \norm{\hat{\theta}_h^{(s)} - \theta_h} &\leq \lambda\Delta G_s^\U + \sigma \sqrt{\frac{3((1-\alpha)m + d)}{m_h^\U}} + 2\alpha_h^\U C\sigma \sqrt{d} \\
                                         &\leq (\lambda G_s^\U + \frac{3}{r_1})\Delta + 2\alpha_h^\U C\sigma \sqrt{d}
\end{align*}
Taking the min of the two terms, the proof is now complete.

\subsubsection{Proof of Lemma~\ref{lemma:a7}}
  Arguing similar to the proof of Lemma A.7 of \cite{lu2016statistical}, we can begin at the result

  \begin{align*}
  m_{gh}^{\U(s+1)} \le m_{gh}^{(s+1)} \le \sum_{i\in T_g^*} \mathbbm{1}{\left\{\frac{\epsilon^2}{4}\norm{\theta_g - \theta_h}^2 \le \langle \tau_i, \theta_h - \theta_g \rangle \right\}} + \sum_{i\in T_g^*} \frac{16}{\epsilon^4\Delta^4}(\tau_i^T(\Delta_h - \Delta_g))^2
  \end{align*}
  Note, $\Delta_h = \hat{\theta}_h^{(s)} - \theta_h$ for $h \in [k]$. By the Lemma~\ref{lem:sub_gauss_A.5}, we can bound the first term in the RHS as
  \begin{align*}
    m_g^*\exp(-\frac{\epsilon^4\Delta^2}{32\sigma ^2}) + \sqrt{5m_g^*\log ((1-\alpha)m)}
  \end{align*}
  Using Lemma~\ref{lem:sub_gauss_A.2} we can bound the second term as well as
  \begin{align*}
  \sum_{i\in T_g^*} \frac{16}{\epsilon^4\Delta^4} (\tau_i^T(\Delta_h - \Delta_g))^2 \leq \frac{96(m_g^* + d)\sigma ^2}{\epsilon^4\Delta^4}\norm{\Delta_g - \Delta_h}^2
  \end{align*}
  Together with the bound $\norm{\Delta_g - \Delta_h}^2 \leq 4\Lambda_s^2\Delta^2$, we get
  \begin{align*}
    m_{gh}^{\U(s+1)} \le m_g^*\exp(-\frac{\epsilon^4\Delta^2}{32\sigma ^2}) + \sqrt{5m_g^*\log ((1-\alpha)m)} + \frac{384(m_g^* + d)\sigma ^2}{\epsilon^4\Delta^2}\Lambda_s^2
  \end{align*}
  We can take the max over all clusters to get
  \begin{align*}
    \max_{g\in [k]} \sum_{h\neq g} \frac{m_{gh}^{\U(s+1)}}{m_g^*} \leq k\exp(-\frac{\epsilon^4\Delta^2}{32\sigma ^2}) + k\frac{\sqrt{5\log ((1-\alpha)m)}}{m_g^*} + \frac{384\sigma^2}{\epsilon^4r_1^2}\Lambda_s^2
  \end{align*}

  Since $\Lambda_s \leq \frac{1}{2}$ and $r_1 \ge 20\epsilon^2$, when $\gamma_1(1-\alpha) \ge 32\log ((1-\alpha)m)$ we have
  \begin{align*}
    \min_{g \in [k]} \frac{m_{gg}^\U}{m_g^*} = 1 - \max_{g\in [k]} \sum_{h\neq g} \frac{m_{gh}^{\U(s+1)}}{m_g^*} \geq \frac{1}{2}
  \end{align*}
  Thus, for some $h \in [k]$,
  \begin{align*}
    (1-\alpha_h^\U)m_{h}^{\U(s+1)} \geq m_{hh}^{\U(s+1)} \geq \frac{1}{2}m_h^*  \geq \frac{1}{2}\gamma_1 (1-\alpha)m
  \end{align*}

  We apply this to get
  \begin{align*}
    \max_{h\in [k]} \sum_{g\neq h} \frac{m_{gh}^{\U(s+1)}}{(1-\alpha_h^\U)m_h^{\U(s+1)}} \leq \frac{2}{\gamma_1}\exp(-\frac{\epsilon^4\Delta^2}{32\sigma^2}) + \sqrt{\frac{5k\log ((1-\alpha)m)}{\gamma_1^2(1-\alpha)m}} + \frac{768}{\epsilon^4r_1^2}\Lambda_s^2
  \end{align*}
  These, two bounds give us
  \begin{align*}
    G_{s+1}^\U \leq \frac{2}{\gamma_1}\exp(-\frac{\epsilon^4\Delta^2}{32\sigma^2}) + \sqrt{\frac{5k\log ((1-\alpha)m)}{\gamma_1^2(1-\alpha)m}} + \frac{768}{\epsilon^4r_1^2}\Lambda_s^2
  \end{align*}

   Assuming $\epsilon^4\gamma_1\Delta^2/\sigma^2 \ge r_1^2\epsilon^4 \ge 36$ we get the result
  \begin{align*}
    G_{s+1}^\U \leq \frac{2}{\epsilon^4r_1^2} + \left(\frac{28}{\epsilon^2r_1}\Lambda_s\right)^2 + \sqrt{\frac{5k\log((1-\alpha)m)}{\gamma_1^2(1-\alpha)m}}
  \end{align*}

\subsubsection{Proof of final results}
By Assumption~\ref{asm:k_mix}, Lemma~\ref{lemma:a6} parallels Lemma A.6 in \cite{lu2016statistical} with an additional term on the order of a constant, $c$, where $\alpha' \le \frac{c}{\sqrt{d}}$. Setting $\epsilon = \frac{12}{\sqrt{r}}$, $C_1 \ge 500$, and under the assumption that $r_1 \ge 160\sqrt{k}$ we can characterize $\delta$. For example, with $\frac{2C\sigma}{\Delta} \le \frac{1}{4}$, $\delta = \frac{1}{2}$ suffices to guarantee $G_s^\U \le 0.35$ and $\Lambda_s \le \frac{1}{2} - \frac{c}{2}$ for all $s$. Thus, given an analogous recursive relationship between the two lemmas, we know the conditions for the lemmas will hold for any step $s$ as long as
  \begin{align*}
    \Lambda_0 \leq \frac{1}{2} - \frac{4}{\sqrt{r_1}} - \delta c
  \end{align*}
  or equivalently,
  \begin{align*}
    G_0^\U \leq (\frac{1}{2} - \frac{6}{\sqrt{r_1}} - \delta c)\frac{1}{\lambda}
  \end{align*}
  holds for some constant $\delta$ to combat the additional offset produced by the extra term. Along with Assumptions~\ref{asm:k_mix} with $C_1 = 16$, we can simplify Lemma~\ref{lemma:a6}, as
  \begin{align*}
    \Lambda_s \le \frac{3}{r_1} + \frac{3}{r_1}\sqrt{kG_s^\U} + G_s^\U + 2Cc \frac{C_3}{r_1} \le \frac{1}{2} + G_s^\U + \frac{CC_3c}{8}
  \end{align*}
  Combining this with Lemma~\ref{lemma:a7} with some constant $C_2$ and $C_4 = \frac{CC_3c}{8}$, we get
  \begin{align*}
    G_{s+1}^\U &\leq \frac{C_2}{r_1^2} + \frac{C_2}{r_1^2}(\frac{1}{4} + G_s^\U + (G_s^{\U})^2 + \frac{C_4}{2} + C_4G_s^\U + C_4^2) + \sqrt{\frac{5k\log((1-\alpha)m)}{\gamma_1^2(1-\alpha)m}} \\
               &\leq \frac{2C_2}{r_1^2} + \frac{2C_2}{r_1^2}G_s^\U + \frac{C_2C_4}{r_1^2}(G_s^\U + C_4) + \sqrt{\frac{5k\log((1-\alpha)m)}{\gamma_1^2(1-\alpha)m}}
  \end{align*}
  which upon further simplification yields the result.

\paragraph{Error floor}
From the above result, $G_{s+1}^\U$ satisfy the following inequality
$$
G_{s+1}^\U \leq \frac{c_1}{r_1^2} G_s^\U + \frac{c_2}{r_1^2}+ \sqrt{\frac{5K\log((1-\alpha)m)}{\gamma_1^2(1-\alpha)m}}
$$
For a sufficiently large $C$, we can write,
$$
G_{s+1}^\U \leq \frac{c_1'}{r_1^2}G_s^\U + \frac{c_2'}{r_1^2} + \sqrt{\frac{5K\log((1-\alpha)m)}{\gamma_1^2(1-\alpha)m}}
$$

with $r_1^2 > c_1'$. Let $\varrho_1 := \frac{c_2'}{r_1^2} + \sqrt{\frac{5K\log((1-\alpha)m)}{\gamma_1^2(1-\alpha)m}}$ and $\delta' = \frac{c_1'}{r_1^2}$. We have
$$
G_{s+1}^\U \leq \delta' G_s^\U + \varrho_1
$$
Iterating this for $S$ iterations, where $S$ is a constant, we get
\begin{equation}
\label{eqn:err_floor}
G_S^\U \leq (\delta')^S G_0^\U + \varrho \left( 1+\delta'+\ldots+(\delta')^{S-1} \right ) \leq \frac{1}{2}(\delta')^S + \frac{1}{1-\delta'}\varrho_1
\end{equation}

We now relate $G_s^\U$ to the untrimmed, cluster-wise misclustering rate $G_s$. To upper bound $G_s$, we only need to bound the first term of $G_s$ because the second term of $G_s$ strictly increases after trimming. Assume the first term dominates in $G_s$.
\begin{align*}
  G_s &= \max_{h\in [K]} \frac{\sum_{g\neq h \in [K]}m_{gh}}{(1-\alpha_h)m_h} \\
      &= \max_{h\in [K]} \frac{\sum_{g\neq h \in [K]}m_{gh}^\U + m_h^\mathcal{T}}{(1-\alpha_h)m_h} \\
      &= \max_{h\in [K]} \frac{(1-\alpha_h^\U)m_h^\U}{(1-\alpha_h)m_h} \frac{\sum_{g\neq h \in [K]}m_{gh}^\U}{(1-\alpha_h^\U)m_h^\U} + \frac{m_h^\mathcal{T}}{(1-\alpha_h)m_h} \\
      &\le \max_{h\in [K]} \frac{1}{1-\alpha_h} \left ( (1 - \beta_h)G_s^\U  + \beta_h \right )
\end{align*}

Thus, we can apply our error floor to $G_s$ with an additional term.
\begin{align*}
  G_S \le \Gamma'(\frac{1}{2}(\delta')^S + \frac{1}{1-\delta'}\varrho_1) + \zeta
\end{align*}

Now plug the values of $\delta'$, we get that for $S \geq 2$, the first term in Equation~\ref{eqn:err_floor} is order-wise negligible compared to the second term. Hence, we get $G_S \leq \varrho$ for $S \geq 2$.

\subsection{High dimension: proof of Theorem~\ref{thm:him_dim_main}}

For the $t$-th iteration, suppose that $y_i$ is a good data point with label $\nu_i$. Since $\hat{\nu}_i^{(t)} = \underset{\nu\in\{-1, 1\}}{\mathrm{argmin}} \twonms{\nu y_i - \hat{\theta}^{(t-1)}}^2 $, we have $I \{ \hat{\nu}_i^{(t)} \neq \nu_i \} = I \{ \innerps{\theta^* + \tau_i}{\hat{\theta}^{(t-1)}} \leq 0\}$. Define the following probability

\begin{equation}
\label{eq:define_prob}
A_t := \PP_{\tau\sim\D_\tau}\left\{  \innerps{\theta^* + \tau}{\hat{\theta}^{(t-1)}} \le 0 \right\}.
\end{equation}
We start by analyzing $A_1$. When $t=1$, since $\tau$ is $\zeta$-sub-Gaussian, we have
\[
A_1 = \prob{ \innerps{\tau}{\hat{\theta}^{(0)}} \le -\innerps{\theta^*}{\hat{\theta}^{(0)}} } \le \exp\left(- \frac{\innerps{\theta^*}{\hat{\theta}^{(0)}}^2}{2\zeta^2 \twonms{\hat{\theta}^{(0)}}^2} \right).
\]
By simple algebra, one can check that for any two vectors $\theta$ and $\hat{\theta}$, we have
\begin{equation}\label{eq:linear_alg}
\innerps{\theta}{\frac{\hat{\theta}}{\twonms{\hat{\theta}}}}^2 \ge \twonms{\theta}^2 - \twonms{\theta - \hat{\theta}}^2.
\end{equation}
Combining with initialization of Assumption~\ref{asm:high_dim} we have
\begin{equation}\label{eq:bound_a1}
A_1 \le \exp(-\frac{3}{8\zeta^2} \twonms{\theta^*}^2)
\end{equation}
Due to the SNR condition of Assumption~\ref{asm:high_dim}, we know that $A_1 \le \frac{1}{16}$. For simplicity, let $n_0=\frac{n}{T}$ denote the number of data points used in each iteration. Thus the data points used in the $t$-th iteration are $y_{(t-1)n_0+1}, y_{(t-1)n_0+2}, \ldots, y_{tn_0}$. We use induction to prove the following result: for any $\delta > 0$, suppose that $ n_0 \ge \frac{1}{\alpha}(d + \frac{1}{\alpha}\log(2/\delta)) $, then, with probability at least $1 - 2t\delta $ over the data points used in the first $t$ iterations, we have
\begin{equation}\label{eq:induction}
A_{t+1} \le \frac{1}{8}(A_t + 3\alpha) +  \exp\left( -\frac{\twonms{\theta^*}^2}{2\zeta^2} \right) \le \frac{1}{16}.
\end{equation}

Suppose that~\eqref{eq:induction} holds for $t-1$, then we consider the $t$-th iteration. We partition the $n_0$ data points used in the $t$-th iteration into three parts: $N_{t,0}$ data points which are inliers and have correct estimated labels (i.e., $\hat{\nu}_i^{(t)} = \nu_i$); $N_{t,1}$ data points which are inliers and the estimated labels are wrong; $N_{t,2}$ data points which are outliers.

According to our contamination model and Hoeffding's inequality, we have
\begin{equation}\label{eq:number_outlier}
\prob{N_{t,2} \ge 2\alpha n_0} \le \exp(-2\alpha^2 n_0).
\end{equation}
Conditioned on all the previous iterations and the fact that there are $N_{t,2}$ outliers, we can use Hoeffding's inequality to bound $N_{t,1}$:
\begin{equation}\label{eq:number_wrong_label}
\prob{N_{t,1} \ge (A_t+\alpha) (n_0 - N_{t,2}) \mid N_{t,2}} \le \exp(-2\alpha^2 (n_0 - N_{t,2})).
\end{equation}
Therefore, with probability at least $ 1 - 2\exp(-\alpha^2 n_0) $, we have
\begin{equation}\label{eq:number_good}
N_{t,0} \ge (1 - ( A_t + 3\alpha ))n_0.
\end{equation}
This implies that if $n_0 \ge \frac{1}{\alpha^2}\log(2/\delta)$, then with probability $1-\delta$,~\eqref{eq:number_good} holds.

The next step in the algorithm is to use the iterative filtering algorithm subroutine to conduct a robust mean estimation of $\theta^*$. To this end, we first construct $n_0$ inliers: for every $i = (t-1)n_0 + 1, (t-1)n_0 + 2, \ldots, tn_0$, if $y_i$ is an inlier, we let $\tilde{y}_i := \nu_i y_i$; if $y_i$ is an outlier, we draw $\tau_i$ from $\D_\tau$ independently from all other data points, and let $\tilde{y}_i := \theta^* + \tau_i$. Here we note that all the new data points $\tilde{y}_i$ (with $y_i$ being outliers) are \emph{virtual}, i.e., they are only used for the analysis purpose.

Now we have $n_0$ virtual data points $\tilde{y}_i$, $i = (t-1)n_0 + 1, (t-1)n_0 + 2, \ldots, tn_0$ drawn from the inlier distribution with mean $\theta^*$. In the algorithm implementation, we have $\hat{\nu}_i^{(t)}y_i$, $i = (t-1)n_0 + 1, (t-1)n_0 + 2, \ldots, tn_0$. In fact, if $y_i$ is an inlier and has correct label, we have $\tilde{y}_i = \hat{\nu}_i^{(t)}y_i$. Thus, the set of data points used in the algorithm $\{\hat{\nu}_i^{(t)}y_i\}$ can be considered as a corrupted sample from $\{\tilde{y}_i\}$. Conditioned on the event in~\eqref{eq:number_good}, we know that with probability at least $ 1 - 2\exp(-\alpha^2 n_0) $, the fraction of corrupted data (including outliers and inliers with wrong labels) is at most $ A_t + 3\alpha $. According to~\cite{diakonikolas2017being,steinhardt2017resilience}, the following lemma holds deterministically.

\begin{lemma}\label{lem:robust_high_dim}
~\cite{diakonikolas2017being,steinhardt2017resilience} Suppose that $A_t + 3\alpha \le \frac{1}{4}$. Let $\bar{y} = \frac{1}{n_0}\sum_{i=(t-1)n_0 + 1}^{tn_0} \tilde{y}_i$ and suppose that
\[
\twonm{\frac{1}{n_0}\sum_{i=(t-1)n_0 + 1}^{tn_0} (\tilde{y}_i - \bar{y} )( \tilde{y}_i - \bar{y} )^\top } \le \hat{\sigma}^2.
\]
Let $\hat{\theta}^{(t)}$ be the output of the iterative filtering algorithm. Then, there exists an absolute constant $c$ such that $\twonms{\hat{\theta}^{(t)} - \bar{y}} \le c_0\hat{\sigma}\sqrt{A_t + 3\alpha}$.
\end{lemma}
Using similar derivations as in~\cite{yin2018defending}, by choosing proper value of $\hat{\sigma}$ as a parameter in the iterative filtering algorithm, we know that there exists an absolute constant $c_1$ such that with probability $1-\delta$,
\begin{equation}\label{eq:parameter_est}
\twonms{\hat{\theta}^{(t)} - \theta^*} \le c_1\left( (\sigma+\zeta)\sqrt{A_t + 3\alpha} + \zeta \sqrt{\frac{d + \log(2/\delta)}{n_0}} \right)
\end{equation}
Suppose that $n_0 \ge \frac{1}{\alpha}(d+\log(2/\delta))$. Then we have
\begin{equation}\label{eq:parameter_est2}
\twonms{\hat{\theta}^{(t)} - \theta^*} \le c_2 (\sigma+\zeta)\sqrt{A_t + 3\alpha}.
\end{equation}

Then we proceed to analyze $A_{t+1}$. According to~\eqref{eq:define_prob}, we have
\[
A_{t+1} = \PP_{\tau\sim\D_\tau}\left\{  \innerps{\theta^* + \tau}{\hat{\theta}^{(t)}} \le 0 \right\}
\]
Since $\tau$ is sub-Gaussian, similar to the derivation of $A_1$, we have
\[
A_{t+1} \le \exp\left( -\frac{\twonms{\theta^*}^2 - \twonms{\theta^* - \hat{\theta}^{(t)}}^2}{2\zeta^2}\right) \le \exp\left( -\frac{\twonms{\theta^*}^2}{2\zeta^2} \right)\exp\left( \frac{c_2^2(\sigma+\zeta)^2(A_t + 3\alpha)}{2\zeta^2} \right).
\]
Since we assume $\frac{\sigma}{\zeta} \le \Theta(1)$, we have
\begin{align*}
A_{t+1} \le & \exp\left( -\frac{\twonms{\theta^*}^2}{2\zeta^2} \right) \exp(c_3(A_t + 3\alpha))  \\
\le & \exp\left( -\frac{\twonms{\theta^*}^2}{2\zeta^2} \right) (1+ e^{c_3} (A_t + 3\alpha) )  \\
\le & \exp\left( c_3 -\frac{\twonms{\theta^*}^2}{2\zeta^2} \right) (A_t + 3\alpha) +  \exp\left( -\frac{\twonms{\theta^*}^2}{2\zeta^2} \right).
\end{align*}
Thus, as long as $\frac{\twonms{\theta^*}^2}{2\zeta^2}$ is greater than or equal to a constant that is large enough, and we can guarantee that $\exp( c_3 -\frac{\twonms{\theta^*}^2}{2\zeta^2} ) \le \frac{1}{8}$ and $\exp( -\frac{\twonms{\theta^*}^2}{2\zeta^2} ) \le \frac{1}{32}$, we have
\[
A_{t+1} \le \frac{1}{8}(A_t + 3\alpha) +  \exp\left( -\frac{\twonms{\theta^*}^2}{2\zeta^2} \right) \le \frac{1}{16}.
\]
Combining the probabilistic arguments~\eqref{eq:number_good} and~\eqref{eq:parameter_est} and by union bound, we know that conditioned on the first $t-1$ iterations, and the fact that~\eqref{eq:induction} holds for $t-1$, and $n_0 \ge \frac{1}{\alpha}(d+\frac{1}{\alpha}\log(2/\delta))$, with probability at least $1-2\delta$ over the data points in the $t$-th iteration, we have~\eqref{eq:induction} holds for $t$.

Thus, we have proved~\eqref{eq:induction}. The final results can be obtained by iterating~\eqref{eq:induction}.

\section{Guarantees of stage I and III of the modular algorithm}

\subsection{Guarantees for stage-I:}
 \label{sec:guarantee}
 
\subsubsection{ERM computation}

 We now show closeness guarantees of the ERMs, $\hat{\vecw}^{(i)}$ to its true risk minimizer, $\vecw_k^*$ for some $k \in [K]$. The closeness results will be required for the provable guarantees of the threshold based clustering algorithm described in Section~\ref{sec:naive_cluster}.

Suppose machine $i \in \mathcal{C}_k$. The goal is to prove an upper bound on $\norms{\vecw^{(i)}-\vecw^*_k}$. We now present the result for completeness.

\begin{theorem}
\label{thm:shai_closeness}\cite{shai}
Under Assumptions~\ref{ass:second_lipschitz},~\ref{ass:strongly_convex} and~\ref{ass:second_smooth}, with probability at least $1-\delta$, we have
\begin{align*}
F_k(\hat{\vecw}^{(i)}) -F(\vecw_k^*) \leq \mathcal{O} ( \frac{G_1^2 L_1 \log(1/\delta)}{\lambda^2 n})
\end{align*}
\end{theorem}
Using strong convexity (Assumption~\ref{ass:strongly_convex}), with probability exceeding $1-\delta$, we obtain
\begin{align*}
\norms{\hat{\vecw}^{(i)}-\vecw^*_k}^2 \leq \mathcal{O} ( \frac{G_1^2 L_1 \log(1/\delta)}{\lambda^3 n}).
\end{align*}

\subsubsection{Online-to-batch conversion}

 Here the $i$-th compute node runs an \textit{online-to-batch conversion} routine to obtain $\bar{\vecw}^{(i)}$. Suppose the loss function $f(\vecw,.)$ is convex and $G_1$ Lipschitz with respect to $\vecw$, and $\vecw \in \mathcal{W}$ is bounded by $D_1$. \cite{shai} along with Assuption~\ref{ass:strongly_convex} shows that, with probability greater than or equal to $1-\delta$, we have
 \begin{align*}
 {\bar{\vecw}^{(i)}-\vecw^*_k}^2 \leq \mathcal{O} (\frac{D_1^2 G_1^2 \log(1/\delta)}{\lambda n}).
 \end{align*}
From the above results, computing the ERM directly and performing an online optimization over $n$ episodes are order-wise identical. All the closeness results for ERMs holds only for non-Byzantine machines.

\subsection{Stage III-robust distributed optimization}
\label{sec:robust_algo}

Note that, since we have $\alpha m$ Byzantine machines and since we cannot control the clustering of Byzantine machines, the clustering results of this section are not perfect. We let the third phase of the algorithm, i.e., the robust distributed optimization to take care of this.

In the third stage of our algorithm, we implement robust distributed optimization algorithms to learn models for every cluster. After the robust clustering step, we obtain the clustering results $\hat{\mathcal{C}}_1, \hat{\mathcal{C}}_2, \ldots, \hat{\mathcal{C}}_K $. In the adversarial setting, the clustering result is not guaranteed to be perfect. Without loss of generality, we assume that more than half of the worker machines in $\hat{\mathcal{C}}_k $ are from true cluster $\mathcal{C}_k$ for every $k\in[K]$. We denote the fraction of worker machines in $\hat{\mathcal{C}}_k$ that do not belong to $\mathcal{C}_k$ by $\hat{\alpha}_k$ and $\max_{k \in [K]}\hat{\alpha}_k <0.5$. The goal of this stage of the algorithm is to learn a model for every cluster, by jointly using all the machines in $\hat{\mathcal{C}}_k$.

To do this, we use the recently developed Byzantine-robust distributed learning algorithms. These algorithms usually take the following steps: in every iteration, the master machine sends the model parameter to the worker machines; the worker machines compute the gradient of their loss functions with respect to the model parameter; the master machine conducts a robust estimation of the gradients collected from all the worker machines and run a gradient descent step. Here, the robust estimation of gradients is a subroutine of the algorithm, and there are a few choices that we can consider. Some examples are median, trimmed mean, and high dimensional robust estimation algorithms such as iterative filtering. The statistical error rates of robust distributed gradient descent with median and trimmed mean subroutine have been analyzed by~\cite{yin2018byzantine}, and the error rates of the robust distributed gradient descent with iterative filtering has been analyzed by~\cite{yin2018defending}. Without loss of generality, we analyze the $k$th cluster, where $k \in [K]$ and assume $M:=|\hat{\mathcal{C}}_k|$.

We will now state the convergence result of trimmed mean and iterative filtering based robust distributed algorithm. Note that, the number of Byzantine nodes in cluster $\mathcal{C}_k$ is at most $\alpha m$, since in the worst case all the Byzantine nodes will be wrongly clustered together in $\mathcal{C}_k$. Thus, the fraction of Byzantine nodes in this cluster will be at most $\hat{\alpha}_k=\frac{\alpha m}{M}$. Let $w^T$ and $w^0$ are the $T$-th iterate and the initial value of the optimization algorithm respectively and $w^*$ is the global minima. We have the following guarantee.

\begin{theorem}\label{thm:main_trim}~\cite{yin2018byzantine,yin2018defending}
 Suppose that Assumptions~ \ref{asm:partial}, \ref{ass:strongly_convex} and~\ref{ass:second_smooth}  hold,  and $\hat{\alpha}_k  \le \frac{1}{2} - \epsilon$ for some $\epsilon>0$.
With constant step-size of $1/L_1$ and with probability at least $1-\mathcal{O}(\frac{d}{(1+nM)^d})$, after $T$ iterations, we have $
\norms{\vecw^T - \vecw^*} \leq (1-\frac{\lambda_F}{L_1 + \lambda_F})^T\norms{ \vecw^0 - \vecw^* } + \frac{2}{\lambda_F}\Delta'$, where, $\Delta' := \widetilde{\bigo} (\frac{\hat{\alpha}_k d }{\sqrt{n}} + \frac{d}{\sqrt{nM}})$ for trimmed mean and $\Delta':= \widetilde{\bigo} ( \frac{\sqrt{\hat{\alpha}_k}}{\sqrt{n}} + \frac{\sqrt{d}}{\sqrt{nM}} )$ for iterative filtering.
\end{theorem}
\begin{remark}
We can relax Assumption~\ref{ass:strongly_convex} and obtain
\begin{align*}
    F_i(w^T)-F_i(w^*) \leq \widetilde{\bigo} (\frac{\hat{\alpha}_k d }{\sqrt{n}} + \frac{d}{\sqrt{nM}})
\end{align*}
with high probability for the trimmed mean algorithm. Similarly for the iterative filtering, we have
\begin{align*}
    F_i(w^T)-F_i(w^*) \leq \widetilde{\bigo} ( \frac{\sqrt{\hat{\alpha}_k}}{\sqrt{n}} + \frac{\sqrt{d}}{\sqrt{nM}} )
\end{align*}
with high probability.
\end{remark}
\begin{remark}
By running $T \ge \frac{L_1 + \lambda_F}{\lambda_F} \log( \frac{\lambda_F}{2\Delta'}\norms{\vecw^0 - \vecw^*} )$ parallel iterations, we can obtain a solution $ \vecw^T$ satisfying $\twonms{\vecw^T - \vecw^*} \le \mathcal{O}(\Delta') $. Also, as shown by \cite{yin2018defending} the term $\frac{\sqrt{d}}{\sqrt{nM}}$ is unavoidable, and hence the error rate for iterative trimmed means is order optimal in dimension.
\end{remark}

\section{Proof of Theorem~\ref{thm:main_result_2}}
\label{sec:main_result_2_proof}
The proof of the theorem comes directly via combining Theorem~\ref{theorem:k_mix} and \ref{thm:main_trim}. Let $\tilde{\alpha}_i = \left ( \frac{\varrho M_i + \alpha m}{M_i + \alpha m} \right )$. Note that, since $G_S$ denotes fraction of non-Byzantine machines that are mis-clustered, $\tilde{\alpha}_i$ denotes the worst case fraction of Byzantine machines for $i$-th  cluster. Assuming $\max_{i \in [K]} \tilde{\alpha}_i < \frac{1}{2}$ and invoking Theorem~\ref{thm:main_trim} yields the result.

\section{Technical lemmas}
\label{sec:tech_lem}

We now list a few technical lemmas. These lemmas (along with the proofs) appear in the Appendix of \cite{lu2016statistical}, and typically follow from the concentration phenomenon of sub-gaussian random variables. For the sake of completeness of the arguments made in the proofs of Theorem~\ref{thm:2cluster} and \ref{theorem:k_mix}, we are re-writing the results here without proofs.

In the setup of Section~\ref{sec:symmetric_cluster}, suppose we have $\tau_1,\ldots,\tau_t$ such that $\tau_i \sim \mathcal{N}(0,\sigma^2 I_d)$ for all $i \in [t]$ and $\tau_i$'s are independent.

\begin{lemma}
\label{lem:gauss_7.1}
Let $S \subset [t]$ with $\tau_S = \sum_{i \in S}\tau_i$. We have,
\begin{equation}
\norms{\tau_S} \leq \sigma \sqrt{2(n+9d)|S|} \nonumber
\end{equation}
with probability at least $1-\exp(-0.1t)$.
\end{lemma}

\begin{lemma}
\label{lem:gauss_7.2}
Let $\bar{\tau}=\frac{1}{t}\sum_{i=1}^{t}\tau_i$. We have, $\langle \bar{\tau},\theta^* \rangle \geq -\frac{\norms{\theta^*}^2}{\sqrt{t}}$ and $\norms{\bar{\tau}}^2 \leq \frac{3d\sigma^2}{t}+ \frac{\norms{\theta^*}^2}{t}$ with probability at least $1-2\exp(-\frac{\norms{\theta^*}^2}{3\sigma^2})$
\end{lemma}

\begin{lemma}
\label{lem:gauss_7.3}
Consider the matrix $\sum_{i=1}^{t}\tau_i \tau_i^T$. The maximum eigenvalue of the matrix is upper bounded by $1.62(n+4d)\sigma^2$ with probability greater than $1-\exp(-0.1t)$.
\end{lemma}

 We now assume that $\tau_i$ are independent sub-gaussian with zero mean and with parameter $\sigma^2$. We have the following results:

 \begin{lemma}
\label{lem:sub_gauss_A.1}
For $S \subset [t]$, $\tau_S = \sum_{i \in S}\tau_i$ satisfy,
$$
\norms{\tau_S} \leq  \sqrt{\sigma(3(t+d)|S|}
$$
with probability at least $1-\exp(0.3t)$.
\end{lemma}

\begin{lemma}
\label{lem:sub_gauss_A.2}
$\lambda_{max} \bigg (\sum_{i=1}^{t} \tau_i \tau_i^T \bigg ) \leq 6\sigma^2(t+d)$ \\
with probability greater than $1-\exp(-0.5t)$.
\end{lemma}

We now follow the notations of Section~\ref{sec:general_sub_gaussian} for the remaining couple of results.

\begin{lemma}
\label{lem:sub_gauss_A.4}
Let $\tau_{T^*_h}=\sum_{i \in T^*_h}\tau_i$. Then, $\norms{\tau_{T^*_h}} \leq 3\sigma\sqrt{(d + \log m)|T^*_h|}$ for all $h \in [K]$ with probability at least $1-\frac{1}{m^3}$.
\end{lemma}

\begin{lemma}
\label{lem:sub_gauss_A.5}
For fixed $x_1,\ldots,x_k \in \R^d$ and $b >0$,
$$
\sum_{i \in T^*_g} 1\{b\norms{x_h -x_g}^2 \leq \langle \tau_i,\norms{x_h -x_g} \rangle \} \leq n^*_g \exp (-\frac{b^2\Delta^2}{2\sigma^2}) + \sqrt{5n^*_g \log m}
$$
for all $g \neq h$ with probability at least $1-\frac{1}{m^3}$.
\end{lemma}

\end{document}